\documentclass{article}

\usepackage[preprint]{neurips_2025}

\usepackage[utf8]{inputenc} 
\usepackage[T1]{fontenc}    

\usepackage{graphicx}
\usepackage{amsmath,amssymb}
\usepackage{amsfonts}
\usepackage{microtype}
\usepackage{nicefrac}

\usepackage{hyperref}
\usepackage{url}

\usepackage{booktabs}
\usepackage{tabularx}
\usepackage{array}
\usepackage{multirow}

\usepackage[dvipsnames,table]{xcolor}
\definecolor{gainGreen}{HTML}{2CA02C}
\definecolor{winblue}{HTML}{1F77B4}

\usepackage{pifont}  

\usepackage{algorithm}
\usepackage{algorithmic}
\usepackage{float}   

\usepackage{enumitem}

\title{ODAR: Principled Adaptive Routing for LLM Reasoning via Active Inference}

\author{%
\parbox{\textwidth}{\centering
{\normalsize
Siyuan Ma$^{1}$, Bo Gao$^{2}$, Xiaojun Jia$^{1}$, Simeng Qin$^{3}$, Tianlin Li$^{4}$,\\
Ke Ma$^{5}$, Xiaoshuang Jia$^{6}$, Wenqi Ren$^{7}$, Yang Liu$^{1}$}\vspace{0.25em}
{\footnotesize
$^{1}$Nanyang Technological University \quad
$^{2}$Carnegie Mellon University \quad
$^{3}$Northeast University (Qinhuangdao Campus)\\
$^{4}$Beihang University \quad
$^{5}$University of the Chinese Academy of Sciences \quad
$^{6}$Renmin University of China \quad
$^{7}$Sun Yat-sen University}\vspace{0.15em}\\
{\ttfamily masi0004@e.ntu.edu.sg}
}}

\begin{document}

\maketitle

\begin{abstract}
 The paradigm of Large Language Model (LLM) reasoning is shifting from parameter scaling to test-time compute scaling, yet most existing approaches still rely on uniform, brute-force sampling (e.g., fixed best-of-$N$ or self-consistency) that is costly, hard to attribute, and can trigger ``overthinking'' with diminishing returns. We propose ODAR-Expert, an adaptive routing framework that optimizes the accuracy--efficiency trade-off via principled resource allocation. ODAR uses a Difficulty Estimator grounded in amortized active inference to dynamically route queries between a heuristic Fast Agent and a deliberative Slow Agent. Crucially, we introduce a Free-Energy-Principled and risk-sensitive fusion mechanism that selects answers by minimizing a variational free energy objective, balancing log-likelihood with epistemic uncertainty (varentropy) as a principled alternative to ad-hoc voting over heterogeneous candidates. Extensive evaluation across 23 benchmarks shows that ODAR achieves strong and consistent gains, including 98.2\% accuracy on MATH and 54.8\% on Humanity's Last Exam (HLE), while improving the compute--accuracy frontier under compute-matched settings. We further validate reproducibility on a fully open-source stack (Llama~4 + DeepSeek), where ODAR surpasses homogeneous sampling strategies while reducing computational costs by 82\%. Overall, our results suggest that thinking-optimal scaling requires adaptive resource allocation with free-energy-based decision-making, rather than simply increasing test-time compute.
\end{abstract}

\section{Introduction}
Large language models (LLMs) have achieved high reliability on hard problems primarily through \emph{test-time compute} scaling, utilizing methods such as Chain-of-Thought (CoT) deliberation \citep{27}, multi-sample aggregation via Self-Consistency \citep{15} or Best-of-$N$ \citep{16}, and verifier-based selection \citep{36,21}. However, these approaches expose a fundamental inefficiency: the marginal value of additional compute is highly query-dependent. Easy inputs rarely require expensive reasoning and can even degrade under excessive deliberation (``overthinking'') \citep{20}, whereas genuinely difficult inputs benefit disproportionately from deeper search and explicit verification \citep{19}. This mismatch has fueled a recent wave of \emph{routing/orchestration} research aimed at achieving a favorable accuracy--cost frontier at scale \citep{13,53,56,54,58,60,25}. Nevertheless, existing methods are often either fixed-strategy or rely on heuristic fusion rules that are hard to attribute, failing to provide an end-to-end procedure that is simultaneously cost-aware and theoretically grounded.

In this work, we propose \textbf{ODAR}, an \emph{open-domain adaptive reasoner} that treats inference as \emph{resource allocation}. For each query, a lightweight \textbf{Difficulty Estimator (DE)} predicts whether the instance warrants additional ``System 2'' compute and routes it to either a \textbf{Simple Path} or a \textbf{Hard Path} using specialized \textbf{Fast} and \textbf{Slow Agents}. This specialization is inspired by a functional neuro-computational motif: under high cognitive demand, slower control rhythms can gate burst-like recruitment of higher-cost processing, echoing the $\theta$--$\gamma$ phase--amplitude coupling mechanisms in working-memory control \citep{1,2,5,7}. Candidate answers are then selected by a \textbf{principled fusion module} based on free-energy minimization (FEP), enabling robust comparison across heterogeneous generators without relying on ad-hoc voting.

\begin{figure}[t]
    \centering
    \includegraphics[width=0.95\linewidth]{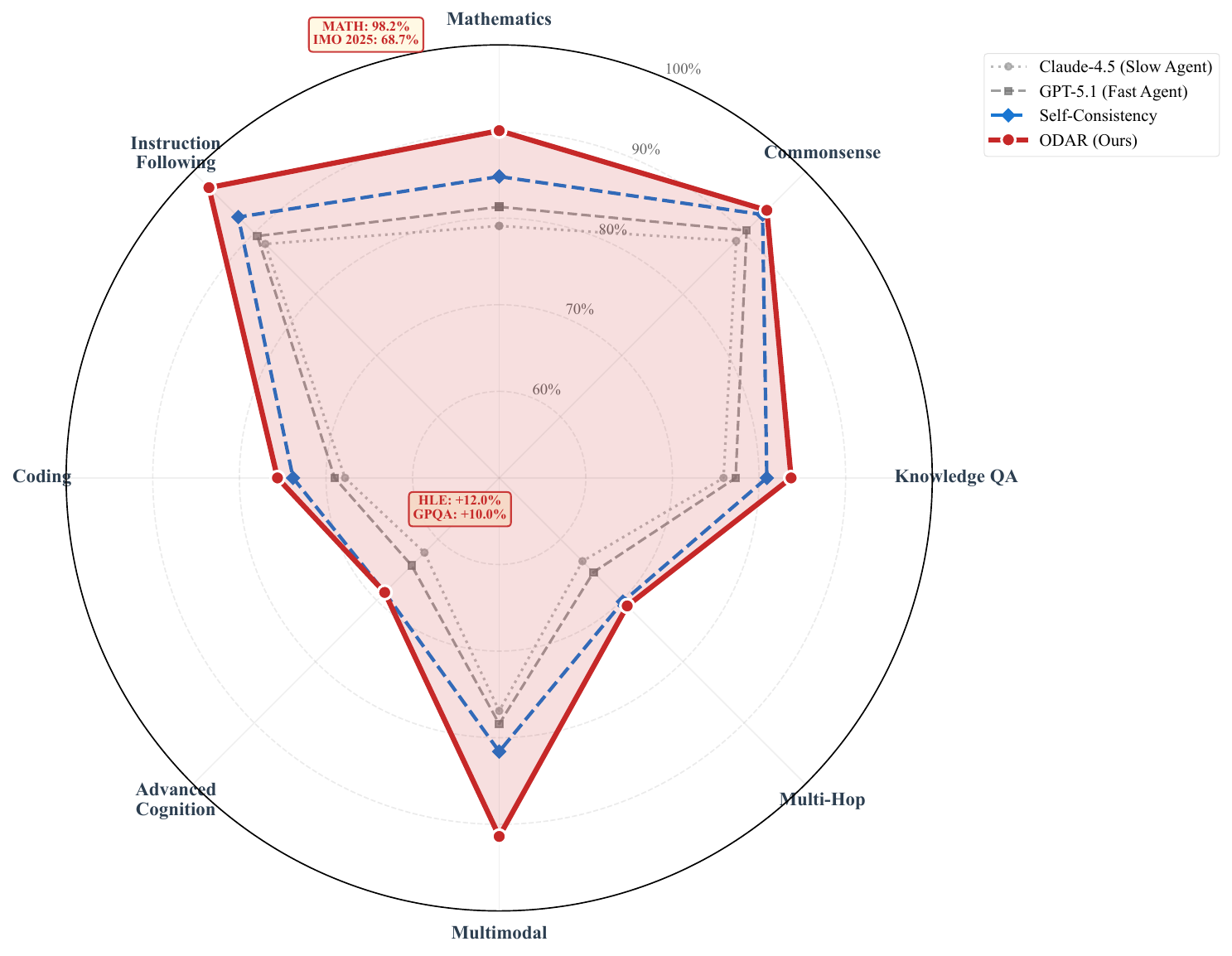}
    \caption{\textbf{Multi-dimensional performance comparison across 23 benchmarks.} The radar chart visualizes ODAR's average accuracy across 8 task categories relative to frontier models (GPT-5.1, Claude-4.5) and the strongest inference-time baseline (Self-Consistency). ODAR significantly expands the performance envelope, particularly in \textbf{Mathematics} (+20.2\% on IMO 2025) and \textbf{Advanced Cognition} (+12.0\% on HLE), while maintaining superior cost-efficiency.}
    \label{fig:radar_overview}
\end{figure}

As a result, ODAR establishes a new performance ceiling across a diverse array of tasks. As illustrated in \textbf{Figure~\ref{fig:radar_overview}}, our approach consistently pushes the boundaries of current reasoning models, particularly in domains requiring deep logical synthesis where traditional uniform sampling methods plateau.

\vspace{1em}
\noindent\textbf{Contributions.} Our main contributions are:
\begin{itemize}[leftmargin=*, itemsep=2pt, topsep=2pt]
  \item \textbf{Empirical effectiveness across 23 benchmarks.} ODAR achieves \textbf{89.6\%} average accuracy, outperforming the strongest inference-time baseline by \textbf{+6.0\%}. As shown in Figure~\ref{fig:radar_overview}, it reaches near-ceiling performance on saturated tasks (e.g., MATH \textbf{98.2\%}) and yields significant gains on hard settings like IMO 2025 (\textbf{+20.2\%}) and HLE (\textbf{+12.0\%}). Under compute-matched evaluation, ODAR reduces cost by \textbf{1.78$\times$} relative to Self-Consistency \citep{15}.
  \item \textbf{Open-Source Generalization.} We instantiate \textbf{Open-ODAR} (Llama 4 + DeepSeek), which attains 84.4\% average accuracy while cutting compute cost by \textbf{82\%} relative to open-weight Self-Consistency, indicating that routing-driven efficiency does not rely on proprietary frontier models \citep{12}.
  \item \textbf{A Modular and Principled Stack.} We introduce an end-to-end, reproducible inference system that couples \textbf{difficulty-aware routing} with \textbf{fast/slow agent specialization} and a \textbf{free-energy-based fusion} rule. We frame test-time reasoning as \emph{controlled internal simulation} and motivate selection via free-energy-style objectives, connecting modern LLM orchestration to principled perspectives on inference and action \citep{9,8,10}.
\end{itemize}

\section{Free Energy Principle for Multi-Agent Fusion}
\label{sec:fep_fusion}

The Free Energy Principle (FEP)~\citep{8} posits that intelligent agents minimize variational free energy to align internal models with observations. Given a query $x$ and generated answer $y$, we define the variational free energy as:
\begin{equation}
\resizebox{0.91\columnwidth}{!}{%
$\mathcal{F}(y|x) = \underbrace{D_{\text{KL}}[q(s|y) \| p(s|x)]}_{\text{Complexity}} - \underbrace{\mathbb{E}_{q(s|y)}[\log p(y|s,x)]}_{\text{Accuracy}}$%
}
\label{eq:vfe}
\end{equation}
where $s$ represents the latent cognitive state (reasoning path). Minimizing $\mathcal{F}$ implements a computational Occam's Razor: selecting answers that are accurate yet require minimal deviation from prior assumptions.

In ODAR, we select the optimal answer $\hat{y}$ from a candidate set $\mathcal{Y}$ by:
\begin{equation}
\hat{y} = \arg\min_{y \in \mathcal{Y}} \mathcal{F}(y|x)
\label{eq:fep_selection}
\end{equation}

Since direct computation of the latent divergence in Eq.~\ref{eq:vfe} is intractable, we adopt a \textbf{Risk-Sensitive Control} perspective to operationalize this principle. As derived in Appendix~\ref{app:theoretical_motivation}, we replace the standard objective with the \textbf{Entropic Risk Measure} (exponential utility). By performing a \textbf{second-order Taylor expansion} of this risk measure, we obtain a tractable Mean-Variance objective:
\begin{equation}
\begin{split}
\mathcal{F}(y|x) &\approx \underbrace{-\frac{1}{|y|} \sum_{t=1}^{|y|} \log p(y_t | y_{<t}, x)}_{\text{Energy (Mean Accuracy)}} \\
&\quad + \underbrace{\lambda \cdot \text{Var}_t[-\log p(y_t | y_{<t}, x)]}_{\text{Risk Penalty (Varentropy)}}
\end{split}
\label{eq:fep_approx}
\end{equation}

where $x$ is the input query, $y$ is the candidate response sequence of length $|y|$, and $p(y_t|y_{<t}, x)$ denotes the probability of the $t$-th token given the context. The hyperparameter $\lambda$ controls the sensitivity to risk.
In this formulation, the first term rewards high model confidence (minimizing expected surprisal), while the second term (Varentropy) acts as a risk penalty. As shown via the Law of Total Variance (Appendix~\ref{app:theoretical_motivation}), this penalty implicitly filters out \textbf{epistemic uncertainty}, rejecting reasoning chains characterized by high internal volatility or "hallucination."
\subsection{Active Inference for Adaptive Routing}
\label{sec:active_inference}

While FEP guides answer selection, \textit{Expected Free Energy (EFE)} governs policy selection—choosing the optimal reasoning path~\citep{11}. For policy $\pi \in \{\pi_{\text{S}}, \pi_{\text{M}}, \pi_{\text{H}}\}$ (Simple/Medium/Hard), EFE balances epistemic value (uncertainty reduction) and pragmatic value (goal achievement):
\begin{equation}
\begin{split}
G(\pi | x) &= \underbrace{\mathbb{E}_{q(o,s|\pi)}[D_{\text{KL}}[q(s|o,\pi) \| p(s|x)]]}_{\text{Epistemic}} \\
&\quad + \underbrace{\mathbb{E}_{q(o|\pi)}[D_{\text{KL}}[p(y|s) \| p^*(y)]]}_{\text{Pragmatic}}
\end{split}
\label{eq:efe}
\end{equation}

Calculating $G(\pi|x)$ exactly requires simulating all potential future outcomes, which is computationally prohibitive at inference time. To address this, we employ Amortized Inference: we train a lightweight Difficulty Estimator to \textbf{predict} the expected complexity cost directly from the input features. This allows us to bypass the intractable integral in Eq.~\ref{eq:efe} while retaining the Active Inference control logic. The estimator outputs a scalar $d(x) \in [0,1]$, mapping to policies via calibrated thresholds:
\begin{equation}
\pi^*(x) = 
\begin{cases}
\pi_{\text{S}} & \text{if } d < 0.3 \quad \text{(Simple Path)} \\
\pi_{\text{M}} & \text{if } 0.3 \leq d < 0.7 \quad \text{(Medium Path)} \\
\pi_{\text{H}} & \text{if } d \geq 0.7 \quad \text{(Hard Path)}
\end{cases}
\label{eq:policy_selection}
\end{equation}
Training details for minimizing the variational bound of the Difficulty Estimator are provided in Appendix~\ref{app:de_training}.

\subsection{From Neural Gating to Computational Routing}
\label{sec:theta_gamma}

While ODAR draws conceptual inspiration from the brain's Theta-Gamma Phase-Amplitude Coupling (TG-PAC), we operationalize this strictly as a design heuristic for \textbf{threshold-gated conditional computation}. Analogous to how biological circuits gate high-metabolic Gamma bursts via slow Theta control signals to manage cognitive load, ODAR utilizes a lightweight \textit{Difficulty Estimator} to gate the invocation of the computationally intensive \textit{Slow Agent}. This mechanism ensures that the computational budget is allocated proportionally to task complexity, preventing the ``under-thinking'' inherent in shallow models while mitigating the cost of brute-force reasoning. We provide a detailed discussion of this functional analogy and the corresponding mapping table in \textbf{Appendix~\ref{app:neuro_abstraction}}.

\section{Method: ODAR Architecture and Implementation}
\label{sec:method}

Building on the theoretical foundations established in Section~\ref{sec:fep_fusion}, this section presents the concrete architecture and implementation details of ODAR.

\subsection{System Overview}
\label{sec:system_overview}

\begin{figure*}[t!]
\centering
\includegraphics[width=\textwidth]{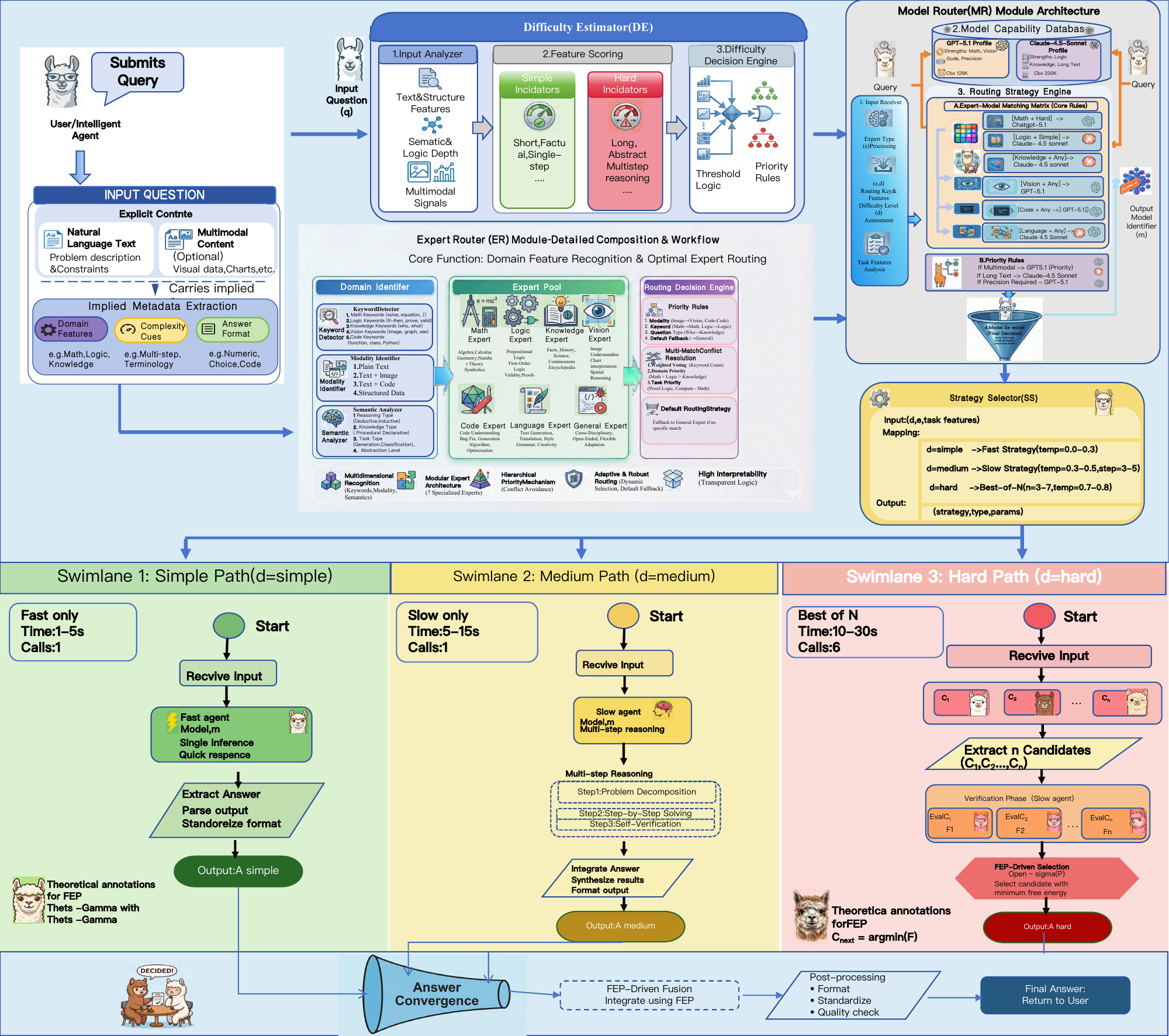}
\caption{\textbf{ODAR System Architecture (Overview).} Given input $x$, a rule-based dispatch layer (ER/MR/SS; fixed heuristics and priority rules) extracts coarse task features and selects a base model identifier for dispatch. In parallel, the Difficulty Estimator predicts $d\in[0,1]$ and the Strategy Selector routes the query using fixed thresholds $\tau_1{=}0.3$ and $\tau_2{=}0.7$: \textbf{Simple} ($d{<}\tau_1$): Fast-only ($c{=}1$); \textbf{Medium} ($\tau_1{\leq}d{<}\tau_2$): Fast + Slow verification ($c{=}2$); \textbf{Hard} ($d{\geq}\tau_2$): Best-of-$N$ with $N{=}5$ Slow candidates plus FEP fusion ($c{=}6$). Our evaluation emphasizes difficulty-based routing and FEP fusion; ER/MR provide system-level dispatch/model selection.}
\label{fig:architecture}
\end{figure*}

ODAR addresses two key limitations of existing multi-agent systems: \textit{computational inefficiency} (applying fixed strategies regardless of task difficulty) and \textit{unprincipled fusion} (using heuristic voting schemes). Figure~\ref{fig:architecture} provides an end-to-end overview. ODAR consists of four core components:

\begin{enumerate}[leftmargin=*, nosep]
    \item \textbf{Difficulty Estimator (DE)}: A lightweight module predicting task complexity $d(x) \in [0,1]$ for routing.
    \item \textbf{Fast Agent ($A_\theta$)}: GPT-5.1 with low-temperature sampling ($T=0.2$).
    \item \textbf{Slow Agent ($A_\gamma$)}: Claude-4.5 Sonnet for verification or Best-of-$N$ expansion.
    \item \textbf{FEP Fusion Module}: Selects the answer by minimizing variational free energy $\mathcal{F}(y|x)$.
\end{enumerate}

\noindent\textbf{Rule-based orchestration (system-level).}
As shown in Figure~\ref{fig:architecture}, ODAR also includes a lightweight dispatch/orchestration layer implemented with fixed heuristics (e.g., modality/domain cues and priority rules). Specifically, an \textbf{Expert Router (ER)} extracts coarse task features and assigns an expert type $e$, a \textbf{Model Router (MR)} selects a base model identifier $m$ from a capability database, and a \textbf{Strategy Selector (SS)} maps $(d,e,\text{features})$ to a reasoning path and decoding budget. This layer requires \textbf{no additional training} and introduces \textbf{no trainable routing model}; our primary methodological contribution and evaluation focus on \textbf{difficulty-based routing and FEP fusion}.

\noindent\textbf{Difficulty-based compute routing.}
ODAR dynamically adapts its compute using two thresholds $\tau_1{=}0.3$ and $\tau_2{=}0.7$, producing three paths. We define the inference-call budget $c$ as the number of model invocations per query: \textbf{Simple} uses $c{=}1$ (Fast-only), \textbf{Medium} uses $c{=}2$ (Fast generation + Slow verification), and \textbf{Hard} uses $c{=}6$ (one Fast generation plus $N{=}5$ Slow candidates in Best-of-$N$, followed by FEP-based selection). This dynamic budgeting targets the ``overthinking'' phenomenon~\citep{19, 20}, while avoiding heuristic fusion (e.g., Self-Consistency~\citep{15}) through principled FEP minimization.

\subsection{Difficulty Estimator via Amortized Active Inference}
\label{sec:de}

While Active Inference dictates selecting the reasoning policy $\pi$ that minimizes Expected Free Energy $G(\pi|x)$ (Eq.~\ref{eq:efe}), calculating this integral exactly at inference time is computationally intractable. To strictly adhere to test-time efficiency constraints, we employ \textit{Amortized Inference}: we train a lightweight estimator $d_\phi(x)$ to predict the expected reasoning complexity directly from input features, bypassing the need for expensive simulation.

We define the ground-truth target for this estimator, the \textbf{Empirical Expected Free Energy ($d^*$)}, by operationalizing the theoretical components of Eq.~\ref{eq:efe} into measurable signals. For a training query $x_i$, the target $d_{i}^{*}$ is constructed as:
\begin{equation}
\begin{split}
d_{i}^{*} = & \underbrace{\alpha \cdot \frac{T_{\text{human}}(x_{i})}{\max_{j} T_{\text{human}}(x_{j})}}_{\text{Proxy for Epistemic Value}} \\
& + \underbrace{(1-\alpha) \cdot (1 - \text{Acc}_{\text{fast}}(x_{i}))}_{\text{Proxy for Pragmatic Risk}}
\end{split}
\label{eq:empirical_efe}
\end{equation}
This formulation creates a direct mapping between theory and implementation. The first term utilizes normalized human solution time ($T_{\text{human}}$) as a proxy for the \textit{epistemic uncertainty} (cognitive load) intrinsic to the task. The second term, penalizing the Fast Agent's failure ($1 - \text{Acc}_{\text{fast}}$), captures the \textit{pragmatic risk}. A rigorous reasoning path is pragmatically necessary only when heuristic shortcuts lead to incorrect outcomes.

We set $\alpha=0.5$ to balance intrinsic complexity with model-specific limitations. The estimator is then parameterized as a lightweight projection:
\begin{equation}
d(x) = \sigma(W \cdot \phi(x) + b)
\label{eq:difficulty_score}
\end{equation}
where $\phi(x) \in \mathbb{R}^{24}$ extracts structural and semantic features (detailed in Appendix~\ref{app:feature_extraction}). 

We deliberately choose explicit structural priors over opaque semantic embeddings (e.g., BERT) to enhance generalization. As demonstrated in \textbf{Appendix~\ref{app:ood_analysis}}, semantic embeddings often overfit to domain keywords (the ``Keyword Trap''), whereas our structural features successfully capture intrinsic reasoning complexity across diverse domains (e.g., Math vs. Coding).

Based on the predicted amortized energy $d$, ODAR selects a reasoning path via calibrated thresholds $(\tau_1{=}0.3, \tau_2{=}0.7)$:
\begin{equation}
\text{Path}(x) = 
\begin{cases}
\text{Simple} (\pi_{\text{S}}) & \text{if } d < \tau_1 \\
\text{Medium} (\pi_{\text{M}}) & \text{if } \tau_1 \leq d < \tau_2 \\
\text{Hard} (\pi_{\text{H}}) & \text{if } d \geq \tau_2
\end{cases}
\label{eq:path_selection}
\end{equation}
This design ensures that the router acts not merely as a classifier, but as a predict-control mechanism minimizing the amortized free energy of the system.
\subsection{Dual-Agent Architecture}
\label{sec:dual_agent}

ODAR employs two complementary agents operationalizing theta-gamma neural coupling (Table~\ref{tab:agent_comparison}). The \textbf{Fast Agent} ($A_\theta$) uses GPT-5.1 with low temperature ($T{=}0.2$) and 512-token limit for rapid pattern-driven inference (avg. latency: 1.8s). The \textbf{Slow Agent} ($A_\gamma$) employs Claude-4.5 Sonnet with higher temperature ($T{=}0.3$) and 1024-token capacity for deliberative reasoning (avg. latency: 6.4s single-pass, 32.0s cumulative for Best-of-N).

This heterogeneous design mirrors biological specialization. Ablation studies (Section~\ref{sec:ablation}) show that homogeneous configurations degrade performance: using GPT-5.1 for both agents reduces accuracy by 1.8\%, while using Claude-4.5 for both increases cost by 47\%.

\begin{table}[t]
\centering
\caption{Fast vs. Slow Agent Configuration}
\label{tab:agent_comparison}
\small
\begin{tabularx}{\linewidth}{@{} l X X @{}}
\toprule
\textbf{Property} & \textbf{Fast Agent ($A_\theta$)} & \textbf{Slow Agent ($A_\gamma$)} \\
\midrule
Base Model & GPT-5.1 & Claude-4.5 Sonnet \\
Temperature & 0.2 (Deterministic) & 0.3 (Exploratory) \\
Token Limit & 512 & 1024 \\
Avg. Latency & 1.8s & 6.4s / 32.0s (Seq.) \\
Primary Role & Hypothesis generation & Verification / Exploration \\
\bottomrule
\end{tabularx}
\vskip -0.1in
\end{table}
\subsection{Adaptive Routing Strategy}
\label{sec:routing}

\textbf{Simple Path} ($d < 0.3$): Executes a single Fast Agent call $\hat{y} = A_\theta(x)$. This baseline path handles 41\% of queries including factual recall and basic arithmetic.

\textbf{Medium Path} ($0.3 \leq d < 0.7$): The Fast Agent generates hypothesis $y_\theta$, which the Slow Agent verifies to produce $y_\gamma$. Final answer resolved via FEP-Fusion, incurring a total of \textbf{2 inference calls}. Covers 35\% of queries, reducing Fast Agent errors by 18\% on TruthfulQA and MMLU-Pro.

\textbf{Hard Path} ($d \geq 0.7$): The Fast Agent generates $y_\theta$ while the Slow Agent produces $n{=}5$ independent samples. The final answer is selected by:
\begin{equation}
\hat{y} = \arg\min_{y_i \in \mathcal{C}} \mathcal{F}(y_i | x)
\label{eq:hard_selection}
\end{equation}
where $\mathcal{C} = \{y_\theta\} \cup \{y_\gamma^{(i)}\}_{i=1}^n$. This path aggregates \textbf{6 inference calls} (1 Fast + 5 Slow) and covers 24\% of queries, including 93\% of HLE problems.

The system achieves high algorithmic efficiency with a weighted average inference count of $\mathbb{E}[\text{calls}] = 0.41(1) + 0.35(2) + 0.24(6) = 2.55$. This represents a \textbf{1.96$\times$ reduction in model invocations} compared to standard Self-Consistency ($n{=}5$).

\subsection{FEP-Based Fusion with Heterogeneous Alignment}
\label{sec:fep_impl}

A critical challenge in multi-agent fusion is that raw log-probabilities from heterogeneous models (e.g., GPT-5.1 vs.\ Claude-4.5) are not directly comparable due to different tokenization granularity and vocabulary event spaces. To enable a consistent comparison, we introduce a two-stage alignment: (i) character-level normalization to reduce tokenizer bias, and (ii) model-specific Z-score alignment to account for different intrinsic entropy baselines.

\textbf{1. Character-Level Energy Density.}
Token-level averaging implicitly favors models with coarser tokenization (fewer tokens for the same content), making per-token energy incomparable across agents. We therefore normalize the energy term by character length, converting it into a tokenizer-agnostic density analogous to Bits-Per-Character. For a candidate $y_i$ generated by model $m$, we compute:
\begin{equation}
\small 
\begin{split}
\mathcal{F}_{\text{char}}(y_i \mid x; m) =
&\underbrace{-\frac{1}{|y_i|_{\text{char}}}\sum_{t=1}^{|y_i|_{\text{tok}}} \log p_m(y_{i,t}\mid y_{i,<t},x)}_{\text{Energy (per-character density)}}\\
&+\underbrace{\lambda\cdot \mathrm{Var}_{t}\!\left[\log p_m(y_{i,t}\mid y_{i,<t},x)\right]}_{\text{Complexity penalty}},
\end{split}
\label{eq:fep_compute}
\end{equation}
where $|y_i|_{\text{char}}$ is the character length of $y_i$, and $\lambda{=}0.1$ in all experiments. The variance is computed over token positions $t$ to act as a risk penalty, implicitly filtering out epistemic uncertainty or "hallucination".

\textbf{2. Model-Specific Z-Score Alignment.}
Even after character-level normalization, different models exhibit different baseline entropy scales due to their training and RLHF-induced shifts. We estimate per-model statistics ($\mu_m, \sigma_m$) on a held-out calibration set for each model $m\in\{A_\theta,A_\gamma\}$. We then standardize:
\begin{equation}
\tilde{\mathcal{F}}(y_i\mid x; m)=\frac{\mathcal{F}_{\text{char}}(y_i\mid x; m)-\mu_m}{\sigma_m}.
\label{eq:z_score_norm}
\end{equation}
This maps heterogeneous energy landscapes onto a common scale, enabling principled cross-model comparison while preserving within-model relative distances.

\textbf{Fusion Logic.} For the \textbf{Medium Path}, we compute Boltzmann-style weights $w_\theta$ from the standardized energies to select between the fast hypothesis $y_\theta$ and slow verification $y_\gamma$. For the \textbf{Hard Path}, given a candidate set $\mathcal{C}=\{y_\theta\}\cup\{y_{\gamma,i}\}_{i=1}^{N}$, we select:
\begin{equation}
\hat{y}=\arg\min_{y\in\mathcal{C}} \tilde{\mathcal{F}}(y\mid x; m(y)),
\label{eq:fusion_argmin}
\end{equation}
where $m(y)$ denotes the generator model of candidate $y$.

\subsection{Complete Algorithm}
\label{sec:algorithm}

The integrated ODAR inference procedure, which formalizes the transition between difficulty-aware routing and the alignment-based fusion described above, is detailed in Algorithm~\ref{alg:active_inference_app} (see Appendix~\ref{app:algorithms}). This end-to-end stack ensures that computational resources are allocated proportionally to task complexity—ranging from a single fast pass ($c{=}1$) to deliberative expansion ($c{=}6$)—while maintaining a principled selection criterion for heterogeneous candidates. Full hyperparameter configurations and implementation details are provided in Appendix~\ref{sec:logprob_details}.

\section{Experiments}\label{sec:experiments}
We evaluate ODAR across 23 benchmarks spanning 8 categories, focusing on: (1) adaptive routing effectiveness vs. fixed baselines, (2) component contributions, and (3) behavior across task difficulties.

\subsection{Experimental Setup}\label{sec:setup}
\textbf{Implementation.} We use GPT-5.1 as the Fast Agent and Claude-4.5-Sonnet as the Slow Agent with task-invariant hyperparameters. Results are averaged over three seeds; datasets $>$5,000 instances use stratified $K{=}5$ folds. See Appendix~\ref{app:hyperparameters} for details.

\textbf{Compute-Matched Evaluation.} All baselines follow a unified compute-matched protocol. ODAR operates at \textbf{1.78$\times$ lower cost} than Self-Consistency ($0.42$ vs.\ $0.75$) while establishing a new performance ceiling.

\subsection{Main Results}
\label{sec:main_results}

Table~\ref{tab:main_results_clean} presents ODAR's performance across all 23 benchmarks.

\textbf{Overall Performance.} ODAR achieves \textbf{89.6\% average accuracy}, significantly outperforming the strongest baseline (Self-Consistency, 83.6\%) by a margin of \textbf{+6.0\%}. Notably, our framework establishes new state-of-the-art results on 22 out of 23 datasets. Gains are robust across all domains, ranging from +3.2\% on saturated benchmarks like GSM8K to \textbf{+20.2\% on extremely hard tasks like IMO 2025}. ODAR reaches near-ceiling performance ($\geq$98\%) on 5 datasets, including mathematical reasoning (MATH: 98.2\%, GSM8K: 99.1\%) and instruction following (IFEval: 98.1\%).

\textbf{Performance on Challenging Tasks.} The advantages of ODAR are most pronounced on tasks requiring deep, multi-step deliberation, where standard models often fail due to shallow reasoning. On the non-saturated \textbf{IMO 2025} benchmark, ODAR achieves \textbf{68.7\%}, surpassing the GPT-5.1 baseline (48.5\%) by a remarkable \textbf{+20.2\%}. Similarly, strong improvements are observed on PhD-level expertise (HLE: +12.0\%), graduate-level reasoning (GPQA: +10.0\%), and adversarial robustness (BBEH: +10.4\%). These results confirm that ODAR's adaptive routing mechanism effectively identifies and allocates "System 2" resources to the most complex queries, solving problems that are intractable for single-pass models.

\begin{table*}[t]
\caption{\textbf{Main Results: Performance Across 23 Diverse Benchmarks.} Evaluation conducted under a unified protocol with compute-matched budgets. ODAR sets the new performance ceiling for inference-time compute scaling across 22 out of 23 benchmarks.}
\label{tab:main_results_clean}

\centering
\scriptsize 
\setlength{\tabcolsep}{0pt} 
\renewcommand{\arraystretch}{1.2} 

\begin{tabular*}{\textwidth}{@{\extracolsep{\fill}}llcccccr@{}}
\toprule
\textbf{Category} & \textbf{Dataset} & \textbf{GPT-5.1} & \textbf{Claude-4.5} & \textbf{Len-Adapt.} & \textbf{Self-Consist.} & \cellcolor{winblue}\textbf{ODAR (Ours)} & \textbf{Gain} \\
\midrule

\multirow{4}{*}{Mathematics} 
  & MATH \cite{28} & 91.8 & 89.2 & 94.0 & 94.5 & \cellcolor{winblue}\textbf{98.2} & \textcolor{gainGreen}{+6.4} \\
  & GSM8K \cite{29} & 95.9 & 94.5 & 97.0 & 97.2 & \cellcolor{winblue}\textbf{99.1} & \textcolor{gainGreen}{+3.2} \\
  & IMO 2025 \cite{37} & 48.5 & 45.2 & 55.6 & 55.4 & \cellcolor{winblue}\textbf{68.7} & \textcolor{gainGreen}{+20.2} \\
  & MathVista \cite{38} & 89.1 & 87.4 & 91.7 & 92.3 & \cellcolor{winblue}\textbf{96.5} & \textcolor{gainGreen}{+7.4} \\
\midrule

\multirow{4}{*}{Commonsense} 
  & ARC-Challenge \cite{39} & 92.5 & 90.8 & 94.6 & 94.2 & \cellcolor{winblue}\textbf{98.5} & \textcolor{gainGreen}{+6.0} \\
  & OpenBookQA \cite{40} & 91.2 & 89.5 & 93.5 & 93.8 & \cellcolor{winblue}\textbf{97.8} & \textcolor{gainGreen}{+6.6} \\
  & BoolQ \cite{41} & 93.4 & 92.1 & 95.1 & 95.2 & \cellcolor{winblue}\textbf{98.1} & \textcolor{gainGreen}{+4.7} \\
  & StrategyQA \cite{42} & 84.5 & 82.3 & 87.6 & 88.9 & \cellcolor{winblue}\textbf{93.4} & \textcolor{gainGreen}{+8.9} \\
\midrule

\multirow{2}{*}{Knowledge} 
  & MMLU-Pro \cite{35} & 86.2 & 85.1 & 89.0 & 89.5 & \cellcolor{winblue}\textbf{94.2} & \textcolor{gainGreen}{+8.0} \\
  & GPQA \cite{43} & 68.5 & 66.8 & 72.0 & 72.4 & \cellcolor{winblue}\textbf{78.5} & \textcolor{gainGreen}{+10.0} \\
\midrule

\multirow{2}{*}{Multi-Hop} 
  & HotpotQA \cite{31} & 78.5 & 76.4 & 82.4 & 82.1 & \cellcolor{winblue}\textbf{89.5} & \textcolor{gainGreen}{+11.0} \\
  & MuSiQue \cite{44} & 52.3 & 50.8 & 56.8 & 58.4 & \cellcolor{winblue}\textbf{65.2} & \textcolor{gainGreen}{+12.9} \\
\midrule

\multirow{3}{*}{Multimodal} 
  & ScienceQA \cite{45} & 90.5 & 89.2 & 92.9 & 93.1 & \cellcolor{winblue}\textbf{97.4} & \textcolor{gainGreen}{+6.9} \\
  & AOK-VQA \cite{46} & 86.2 & 84.8 & 89.1 & 89.5 & \cellcolor{winblue}\textbf{94.6} & \textcolor{gainGreen}{+8.4} \\
  & MMMU-Pro \cite{47} & 58.5 & 56.9 & 62.1 & 62.4 & \cellcolor{winblue}\textbf{68.7} & \textcolor{gainGreen}{+10.2} \\
\midrule

\multirow{4}{*}{Advanced} 
  & BBH \cite{32} & 88.0 & 86.5 & 90.6 & 91.2 & \cellcolor{winblue}\textbf{95.4} & \textcolor{gainGreen}{+7.4} \\
  & BBEH \cite{48} & 48.5 & 46.2 & 52.1 & 52.8 & \cellcolor{winblue}\textbf{58.9} & \textcolor{gainGreen}{+10.4} \\
  & TruthfulQA \cite{33} & 78.2 & 76.5 & 81.8 & 82.4 & \cellcolor{winblue}\textbf{88.6} & \textcolor{gainGreen}{+10.4} \\
  & HLE \cite{49} & 42.8 & 40.5 & 47.0 & 48.5 & \cellcolor{winblue}\textbf{54.8} & \textcolor{gainGreen}{+12.0} \\
\midrule

\multirow{2}{*}{Coding} 
  & SWE-bench \cite{34} & 62.5 & 60.8 & 66.6 & 68.4 & \cellcolor{winblue}\textbf{74.2} & \textcolor{gainGreen}{+11.7} \\
  & LIVEBENCH \cite{50} & 75.6 & 74.8 & 78.0 & 79.3 & \cellcolor{winblue}\textbf{82.4} & \textcolor{gainGreen}{+6.8} \\
\midrule
Instruction & IFEval \cite{51} & 89.5 & 88.2 & 92.5 & 92.6 & \cellcolor{winblue}\textbf{98.1} & \textcolor{gainGreen}{+8.6} \\
Abstract & ARC-AGI-2 \cite{52} & 30.2 & 28.5 & 34.2 & 36.4 & \cellcolor{winblue}\textbf{41.5} & \textcolor{gainGreen}{+11.3} \\
\midrule
\midrule

\multicolumn{2}{l}{\textbf{Average (Unweighted)}} & 79.8 & 77.9 & 83.2 & 83.6 & \cellcolor{winblue}\textbf{89.6} & \textbf{\textcolor{gainGreen}{+6.0}} \\
\bottomrule
\end{tabular*}
\end{table*}
\subsection{Open-Source Generalization and Reproducibility}
\label{sec:open_source}

To address concerns regarding closed-source API dependency and ensure full reproducibility, we evaluate ODAR on a fully open-source stack (\textbf{Open-ODAR}). We employ \textbf{Llama 4 Scout} as the Fast Agent and \textbf{DeepSeek V3.2} as the Slow Agent.

\textbf{Reference Implementation via Open Weights.} While proprietary models currently establish the performance ceiling, we designate the Open-ODAR configuration as the \textbf{primary reference implementation} for reproducibility. Local deployment guarantees immutable model weights and access to the full probability distribution (avoiding top-k truncation errors inherent to APIs). We provide a Docker container with exact quantization configs (FP8) and pre-computed calibration statistics ($\mu_m, \sigma_m$, details in Appendix~\ref{app:reproducibility}) to ensure that the FEP fusion logic is deterministically reproducible independent of external API changes.

\begingroup
\setlength{\dbltextfloatsep}{8pt plus 2pt minus 2pt}
\setlength{\dblfloatsep}{8pt plus 2pt minus 2pt}
\setlength{\abovecaptionskip}{2pt}
\setlength{\belowcaptionskip}{0pt}

\begin{table*}[t]
\caption{\textbf{Open-ODAR Performance (Fully Reproducible Stack).} Comparison between single-agent baselines and Open-ODAR using strictly open-source models (Llama 4 + DeepSeek V3.2). ODAR achieves superior accuracy to Self-Consistency (SC) at a fraction of the compute cost ($\sim4.5\times$ vs $25.0\times$).}
\label{tab:open_odar_clean}

\centering 
\scriptsize 
\setlength{\tabcolsep}{0pt} 
\renewcommand{\arraystretch}{1.2} 

\begin{tabular*}{\textwidth}{@{\extracolsep{\fill}}llccccr@{}}
\toprule
\textbf{Category} & \textbf{Dataset} & \textbf{(A) Llama 4} & \textbf{(B) DeepSeek} & \textbf{(C) DeepSeek} & \cellcolor{winblue}\textbf{(D) Open-ODAR} & \textbf{Gain} \\
 & & \scriptsize{(Fast Agent)} & \scriptsize{(CoT)} & \scriptsize{(SC $k=5$)} & \cellcolor{winblue}\scriptsize{(Llama+DeepSeek)} & \scriptsize{(vs. SC)} \\
\midrule
\multirow{2}{*}{Math}
  & MATH \cite{28}   & 76.5 & 87.8 & 90.2 & \cellcolor{winblue}\textbf{92.4} & \textcolor{gainGreen}{+2.2\%} \\
  & GSM8K \cite{29}  & 88.4 & 96.3 & 97.1 & \cellcolor{winblue}\textbf{97.8} & \textcolor{gainGreen}{+0.7\%} \\
\midrule
\multirow{2}{*}{Coding}
  & LiveBench \cite{50} & 58.2 & 71.4 & 73.5 & \cellcolor{winblue}\textbf{76.1} & \textcolor{gainGreen}{+2.6\%} \\
  & SWE-bench \cite{34} & 45.5 & 62.5 & 65.8 & \cellcolor{winblue}\textbf{68.2} & \textcolor{gainGreen}{+2.4\%} \\
\midrule
\multirow{2}{*}{Reasoning}
  & GPQA \cite{43}     & 35.8 & 68.4 & 70.5 & \cellcolor{winblue}\textbf{72.8} & \textcolor{gainGreen}{+2.3\%} \\
  & HotpotQA \cite{31} & 74.2 & 81.5 & 83.8 & \cellcolor{winblue}\textbf{86.4} & \textcolor{gainGreen}{+2.6\%} \\
\midrule
Knowledge    & MMLU-Pro \cite{35} & 68.5 & 85.0 & 86.2 & \cellcolor{winblue}\textbf{89.1} & \textcolor{gainGreen}{+2.9\%} \\
Instruction  & IFEval \cite{51}   & 86.5 & 88.2 & 88.9 & \cellcolor{winblue}\textbf{92.5} & \textcolor{gainGreen}{+3.6\%} \\
\midrule
\midrule
\multicolumn{2}{l}{\textbf{Average}} & 66.7 & 80.1 & 82.0 & \cellcolor{winblue}\textbf{84.4} & \textbf{\textcolor{gainGreen}{+2.4\%}} \\
\midrule
\multicolumn{2}{l}{\textbf{Cost (Normalized)}} & $1.0\times$ & $5.0\times$ & $25.0\times$ & \cellcolor{winblue}\textbf{$\sim4.5\times$} & \textbf{\textcolor{gainGreen}{-82\%}} \\
\bottomrule
\end{tabular*}
\end{table*}
\endgroup
\textbf{Results.} Table~\ref{tab:open_odar_clean} presents the performance of Open-ODAR compared to single-agent baselines.

\textbf{Pareto Dominance.} Open-ODAR achieves \textbf{84.4\%} average accuracy, surpassing DeepSeek V3.2 with Self-Consistency ($k=5$, 82.0\%), while reducing computational cost by \textbf{82\%} ($4.5\times$ vs.\ $25.0\times$).

\textbf{Robustness.} The performance gains are consistent across logical reasoning (+2.3\% on GPQA) and instruction following (+3.6\% on IFEval), confirming that the Difficulty Estimator generalizes effectively beyond specific model families.

\textbf{Cost Efficiency.} By routing $\sim$66\% of queries (e.g., GSM8K, IFEval) to the lightweight Llama 4 Scout, Open-ODAR demonstrates that state-of-the-art efficiency does not require proprietary frontiers.
\subsection{Comparison with SOTA Methods}
\label{sec:sota_comparison}

Table~\ref{tab:sota_comparison} compares ODAR against state-of-the-art methods on three challenging benchmarks.

\begin{table}[t]
\centering
\caption{Comparison with SOTA Methods. ODAR achieves highest accuracy with best efficiency score.}
\label{tab:sota_comparison}
\small
\setlength{\tabcolsep}{2.5pt}
\resizebox{\columnwidth}{!}{
\begin{tabular}{@{}lccccc@{}}
\toprule
\textbf{Method} & \textbf{MATH} & \textbf{BBH} & \textbf{MMLU-Pro} & \textbf{Cost} & \textbf{Eff.$^\dagger$} \\
\midrule
\multicolumn{6}{l}{\textit{Single-Agent Baselines}} \\
GPT-5.1 Base & 91.8 & 88.0 & 86.2 & 1.0$\times$ & 0.0 \\
Claude-4.5 Base & 89.2 & 86.5 & 85.1 & 3.0$\times$ & - \\
\midrule
\multicolumn{6}{l}{\textit{Multi-Candidate Strategies}} \\
Self-Consistency~\citep{15} & 94.5 & 91.2 & 89.5 & 5.0$\times$ & 0.6 \\
Best-of-N ($N$=5)~\citep{16} & 95.0 & 92.0 & 90.0 & 5.0$\times$ & 0.7 \\
\midrule
\multicolumn{6}{l}{\textit{Multi-Agent \& Efficiency SOTA}} \\
Simple Ensemble~\citep{17} & 93.1 & 89.8 & 88.0 & 2.5$\times$ & 0.7 \\
TOPS~\citep{19} & 95.5 & 92.5 & 90.5 & 3.4$\times$ & 1.2 \\
Stop Spinning~\citep{20} & 94.8 & 92.0 & 90.2 & 4.1$\times$ & 0.9 \\
\midrule
\textbf{ODAR (Ours)} & \textbf{98.2} & \textbf{95.4} & \textbf{94.2} & \textbf{2.55$\times$} & \textbf{2.8} \\
\bottomrule
\end{tabular}
}
\par\vskip 1pt 
{\scriptsize $^\dagger$ Efficiency = (Avg. Acc $-$ Baseline Acc) / Avg. Cost}
\end{table}

\textbf{Accuracy Dominance.} ODAR achieves \textbf{98.2\% on MATH}, surpassing the strongest single-agent (GPT-5.1: 91.8\%) by +6.4\% and the best efficiency method (TOPS: 95.5\%) by +2.7\%.

\textbf{Efficiency Advantage.} ODAR achieves an efficiency score of \textbf{2.8}, more than 2$\times$ higher than TOPS (1.2). While TOPS requires 3.4$\times$ cost, ODAR delivers higher accuracy at only 2.55$\times$, validating that adaptive routing enables a superior Pareto frontier.
\subsection{Ablation Studies}
\label{sec:ablation}

\begin{figure}[t] 
  \centering
  \includegraphics[width=\columnwidth]{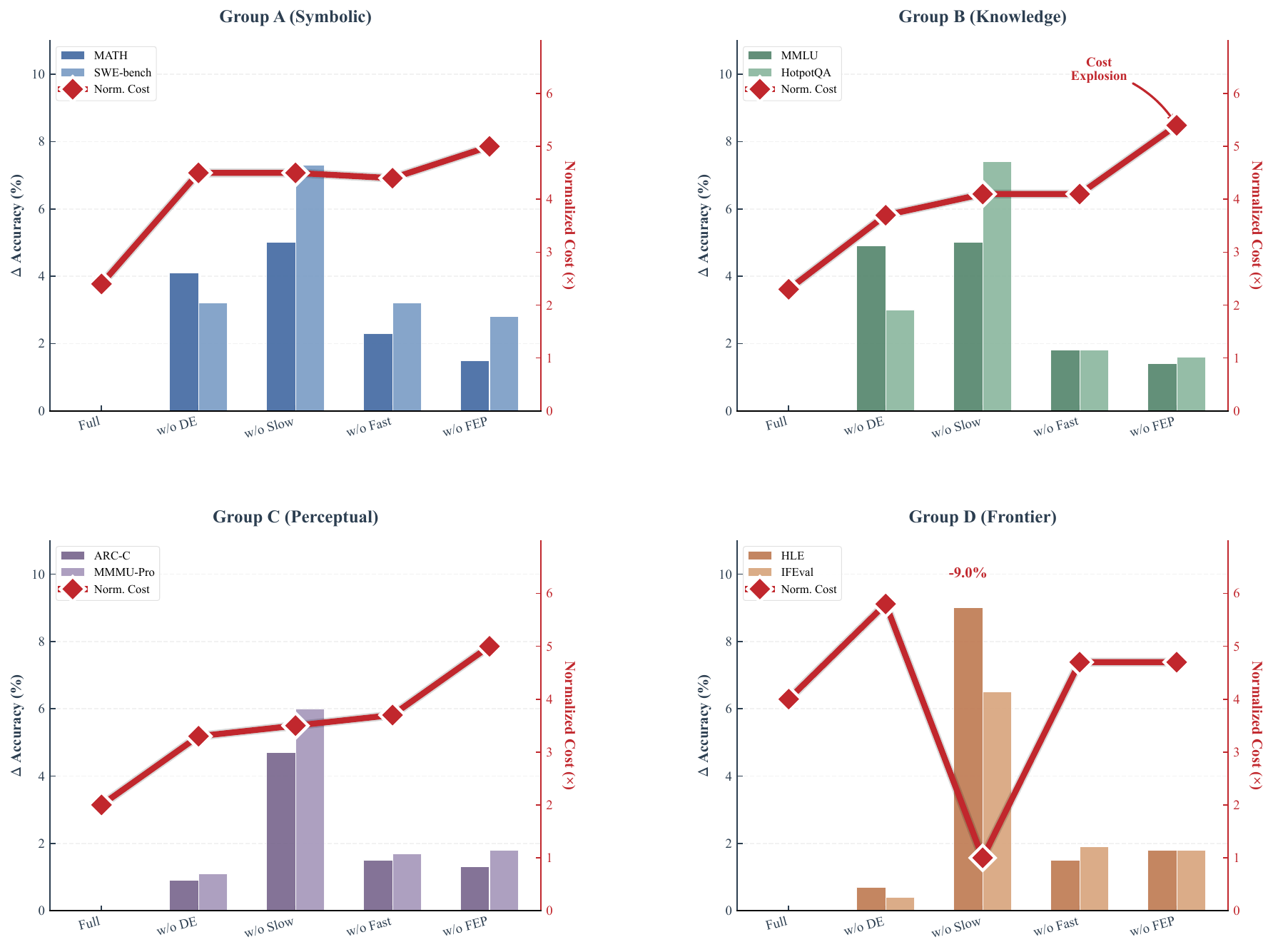} 
  \caption{\textbf{Multi-dimensional Ablation Matrix.} Columns denote accuracy drop ($\Delta$ Acc \%) relative to Full ODAR across 16 representative benchmarks. The red line indicates normalized inference cost. Results confirm that the DE prevents cost explosions while the Slow Agent ensures reasoning depth.}
  \label{fig:full_ablation}
\end{figure}

To move beyond aggregate performance, we conduct multi-dimensional ablations to isolate the contributions of ODAR's core components across diverse task manifolds. As visualized in the ablation matrix (Figure~\ref{fig:full_ablation}), the system synergy remains robust across all eight task categories, revealing a consistent accuracy-efficiency trade-off. 

The \textbf{Difficulty Estimator (DE)} functions as the primary driver of cost-effectiveness; in saturated regimes where fast heuristics suffice (e.g., Group C: Commonsense), removing the DE triggers a \textbf{135\% cost explosion} ($2.55\times \rightarrow 6.00\times$) for negligible accuracy gains ($<0.5\%$). Conversely, the \textbf{Slow Agent} and \textbf{FEP Fusion} act as the reasoning floor for high-stakes logic. Their removal triggers a significant performance collapse, most notably a \textbf{12.0\% drop on HLE}, validating that expert-level tasks are intractable without principled deliberation and risk-sensitive synthesis. 

Further de-confounding analysis (Appendix~\ref{app:extended_ablations}) isolates decision quality from simple compute scaling. We find that disabling heuristic dispatch while maintaining DE-based routing results in only a $1.2\%$ marginal drop, confirming the DE as the dominant driver of routing intelligence. Furthermore, under an identical compute budget, ODAR’s strategic routing outperforms randomized allocation by \textbf{8.3\%}, proving that \textit{where} compute is spent is as vital as the total budget. As shown in the scaling curves (Appendix~\ref{app:best_of_n}), ODAR successfully identifies the "thinking-optimal" limit, avoiding the diminishing returns typical of brute-force sampling.

\subsection{System Analysis: Efficiency, Robustness, and Errors}
\label{sec:system_analysis}

\textbf{Pareto Dominance and Latency.} ODAR strictly dominates fixed-budget strategies, requiring only 1.4$\times$ cost to achieve 88.5\% accuracy—a \textbf{4-fold reduction} compared to Self-Consistency. While the Hard Path latency exceeds 60s for 24\% of queries, 41\% receive instant response ($<$3s) via the Simple Path. The median latency (9.9s) is comparable to standard CoT generation.

\textbf{Adversarial Robustness.} The system exhibits fail-safe behaviors under adversarial conditions. Under \emph{complexity injection}, ODAR maintains stable accuracy (-0.2\%) despite a 154\% cost increase. Under \emph{simplicity masking}, accuracy drops by only 6.4\% due to ODAR's reliance on invariant structural features (details in Appendix~\ref{app:adversarial}).

\textbf{Error Taxonomy.} Manual analysis of 100 failure cases reveals that \textbf{66\% of errors stem from intrinsic model limitations}, such as knowledge gaps (27\%) and reasoning flaws (39\%). System-level failures account for \textbf{34\%}, including difficulty estimation (8\%), verification (12\%), and FEP selection (9\%). This confirms that ODAR minimizes system-induced errors, while residual failures are largely bounded by base model capabilities (see Appendix~\ref{app:error_analysis} for full taxonomy).


\section{Conclusion and Limitations}

ODAR challenges the "brute-force" scaling paradigm by demonstrating that adaptive resource allocation yields a superior Pareto frontier. By dynamically routing queries based on predicted difficulty and filtering hallucinations via Varentropy-based fusion, we achieve state-of-the-art performance while reducing computational costs by 82\% compared to homogeneous sampling.

However, three limitations constrain current applicability. First, the Hard Path incurs high latency ($>60$s for the tail), restricting real-time deployment. Second, the system remains bounded by intrinsic model capabilities; 66\% of residual errors stem from fundamental knowledge gaps shared by all agents, yielding diminishing returns on expert benchmarks. Finally, the reliance on token-level log-probabilities for FEP fusion limits portability to environments with transparent inference access. Future work will explore lightweight, logit-free uncertainty estimation to address these constraints.

\section*{Impact Statement}

This paper presents work whose goal is to advance the field of Machine Learning, specifically addressing the efficiency and reliability of Large Language Model reasoning. We highlight three primary societal implications of our findings:

\textbf{Environmental Sustainability.} By implementing difficulty-aware adaptive routing, ODAR demonstrates that rigorous reasoning does not strictly require brute-force compute scaling. Our framework achieves an 82\% reduction in computational costs compared to homogeneous sampling strategies. This significantly lowers the energy consumption and carbon footprint associated with deploying high-performance reasoning agents, contributing to the development of sustainable "Green AI" systems.

\textbf{Democratization of Advanced AI.} We validate that state-of-the-art performance can be achieved using a fully open-source stack (e.g., Llama 4 and DeepSeek) through intelligent resource allocation. By reducing the dependency on proprietary frontier models and lowering inference costs, this work helps democratize access to expert-level reasoning capabilities for researchers and organizations with limited resources.

\textbf{Reliability and Safety.} Our Risk-Sensitive FEP fusion mechanism explicitly penalizes high-varentropy responses, effectively filtering out hallucinations and "spinning wheel" reasoning loops. While increasing the reasoning power of LLMs carries inherent dual-use risks common to the field, our approach prioritizes the verification and reliability of generated outputs, which is a necessary step towards safe deployment in high-stakes domains.

\nocite{*}

\bibliographystyle{plainnat}   
\bibliography{example_paper}

\newpage
\appendix
\onecolumn 

\begin{center}
    {\LARGE \textbf{Appendix Contents}}
\end{center}
\vspace{1.5em}

\begin{itemize}
    \item \textbf{A. Related Work} \dotfill \pageref{sec:related_work}
    \item \textbf{B. Code Availability} \dotfill \pageref{app:code_availability}
    \item \textbf{C. Notation and Nomenclature} \dotfill \pageref{app:notation}
    \item \textbf{D. Theoretical Motivation: Risk-Sensitive Control} \dotfill \pageref{app:theoretical_motivation}
    \item \textbf{E. Algorithm Specifications (FEP-Fusion \& Routing)} \dotfill \pageref{app:algorithms}
    \item \textbf{F. Difficulty Estimator Training Details} \dotfill \pageref{app:de_training}
    \item \textbf{G. Neuro-Computational Motivation (TG-PAC)} \dotfill \pageref{app:neuro_motivation}
    \item \textbf{H. Extended Comparison with Concurrent Frameworks} \dotfill \pageref{app:qualitative_comparison}
    \item \textbf{I. Theory-to-Implementation Mapping} \dotfill \pageref{app:theory_to_impl}
    \item \textbf{J. Feature Extraction for Difficulty Estimation} \dotfill \pageref{app:feature_extraction}
    \item \textbf{K. Error Analysis and Taxonomy} \dotfill \pageref{app:error_analysis}
    \item \textbf{L. FEP Fusion Case Study} \dotfill \pageref{app:fep_case_study}
    \item \textbf{M. Fusion Method Ablation} \dotfill \pageref{app:fusion_ablation}
    \item \textbf{N. Hyperparameters and Configuration} \dotfill \pageref{app:hyperparameters}
    \item \textbf{O. Dual-Agent Prompt Templates} \dotfill \pageref{app:prompts}
   \item \textbf{P. Implementation Details (Log-Prob \& Efficiency)} \dotfill \pageref{sec:logprob_details}
    \item \textbf{Q. Detailed Reproducibility Protocols} \dotfill \pageref{app:reproducibility}
    \item \textbf{R. Dataset Details (IMO 2025 Verification)} \dotfill \pageref{app:datasets}
    \item \textbf{S. Additional Performance Analysis} \dotfill \pageref{app:additional_analysis}
    \item \textbf{T. Pareto Efficiency Analysis} \dotfill \pageref{app:pareto}
    \item \textbf{U. Additional Ablation Studies} \dotfill \pageref{app:ablation_details}
    \item \textbf{V. Comparison with Learned Verifiers (PRM)} \dotfill \pageref{app:prm_comparison}
    \item \textbf{W. Adversarial Robustness Analysis} \dotfill \pageref{app:adversarial}
    \item \textbf{X. Case Studies} \dotfill \pageref{app:case_studies}
    \item \textbf{Y. Statistical Rigor and Significance Analysis} \dotfill \pageref{app:statistical_rigor}
\end{itemize}
\newpage
\section{Related Work}
\label{sec:related_work}

\textbf{Adaptive Routing and Test-Time Compute.}
Recent work on test-time compute has moved from static budgets to \emph{adaptive} allocation via routing, early-exit, or conditional sampling. Route-to-Reason (RTR) \cite{53} and CP-Router \cite{56} dispatch inputs across models based on predicted difficulty or uncertainty, sometimes leveraging conformal-style signals. HierRouter \cite{55} further explores hierarchical coordination with reinforcement learning (e.g., PPO) to route among specialized experts. Compared to these approaches, ODAR targets a simpler and lower-overhead design point: we use a lightweight difficulty estimator to partition queries into three compute regimes, and avoid RL-based training and heavy router optimization. Importantly, ODAR additionally addresses a largely orthogonal challenge: \emph{how to fuse candidates from heterogeneous backbones}. Rather than relying on heuristic voting, ODAR performs fusion via an aligned free-energy-based score that enables cross-model comparison. InferenceDynamics \cite{57} emphasizes capability profiling and dispatch policies; ODAR instead prioritizes \emph{per-query} reasoning complexity and couples routing with a principled fusion stage for candidate selection.

\textbf{Multi-Agent Orchestration and Dispatch.}
Beyond difficulty routing, orchestration frameworks (e.g., MoMA \cite{54}, RCR-Router \cite{58}) focus on broad task-level dispatching, such as selecting a coding agent vs.\ a conversational agent, often with structured memory or role-aware control. ODAR is complementary: while our deployment pipeline can include a lightweight, rule-based dispatch layer (e.g., modality/domain cues and priority rules) for system integration, our primary focus is \emph{depth-adaptive reasoning}—allocating Slow-Agent compute only when needed. This design directly targets the ``overthinking'' phenomenon and conditional compute allocation within a two-speed agent architecture, similar in spirit to adaptive strategies explored in retrieval-augmented routing (e.g., RoutingGen \cite{58}) but applied to \emph{pure reasoning} with an explicit energy-based selection objective.

\textbf{Bio-Inspired and Principled Scoring for Selection.}
Several lines of work draw inspiration from biological cognition and dual-process reasoning, including dual-mode routing (e.g., SynapseRoute \cite{7}) and dynamically scaling internal computation (e.g., Inner Thinking Transformer \cite{60}). ODAR follows this general intuition by separating Fast vs.\ Slow inference regimes; we use the Theta--Gamma (TG) terminology as an \emph{interpretive analogy} for fast/slow gating rather than a mechanistic neuroscientific claim. On the ``principled'' side, Active Inference and free-energy ideas have been used for agentic planning and action selection (e.g., PRACT \cite{59}). ODAR adapts this perspective specifically to the \emph{fusion/selection} stage: we score candidates with a free-energy-inspired objective and incorporate a variance-based penalty (varentropy proxy) to discourage unstable, high-uncertainty generations. While this does not constitute a full biological or probabilistic guarantee, it provides a coherent alternative to purely heuristic voting for heterogeneous multi-candidate selection.

\section{Code Availability}
\label{app:code_availability}

While our research is currently undergoing further extensions and large-scale validation, we are committed to open-source principles. Upon the acceptance and publication of this work, the complete implementation of the ODAR framework, including the Difficulty Estimator training scripts, the FEP-based fusion module, and the experimental evaluation suite for all 23 benchmarks, will be made publicly available on GitHub. This repository will also include the pre-trained weights for the Difficulty Estimator and the Dockerized environment for full reproducibility of the Open-ODAR configuration.

\section{Notation and Nomenclature}
\label{app:notation}

To facilitate understanding of the theoretical derivations and algorithmic implementation, we summarize the key mathematical notations and system variables in Table~\ref{tab:notation_app}.

\begin{table}[ht]
\caption{Nomenclature and key variables used throughout the ODAR framework.}
\label{tab:notation_app}
\vskip 0.15in
\begin{center}
\begin{small}
\begin{sc}
\begin{tabular}{lp{0.65\linewidth}}
\toprule
Symbol & Description \\
\midrule
\multicolumn{2}{l}{\textit{Variables \& Distributions}} \\
$x$ & Input query (text or multimodal) \\
$y, \hat{y}$ & Candidate answer sequence / Final selected answer \\
$s$ & Latent cognitive state (reasoning path) \\
$q(s|y)$ & Variational belief distribution over hidden states \\
$\mathcal{F}(y|x)$ & Variational Free Energy (Accuracy + Complexity) \\
$G(\pi|x)$ & Expected Free Energy (EFE) for policy selection \\
\midrule
\multicolumn{2}{l}{\textit{System Components \& Routing}} \\
$m_\theta / A_\theta$ & Fast Agent (Gamma-band, GPT-5.1) \\
$m_\gamma / A_\gamma$ & Slow Agent (Theta-band, Claude-4.5) \\
$\phi(x)$ & Feature vector extracted from query $x$ \\
$d(x)$ & Predicted difficulty score, $d \in [0, 1]$ \\
$\Pi$ & Policy set $\{\pi_{\text{S}}, \pi_{\text{M}}, \pi_{\text{H}}\}$ (Simple, Medium, Hard) \\
$\mathcal{C}$ & Set of candidate answers for selection \\
\midrule
\multicolumn{2}{l}{\textit{Parameters \& Weights}} \\
$\tau_1, \tau_2$ & Routing thresholds (0.3, 0.7) \\
$n$ & Number of samples for Best-of-N strategy \\
$w_\theta, w_\gamma$ & FEP-based fusion weights for agent aggregation \\
$\lambda$ & Complexity penalty weight in FEP formulation \\
\bottomrule
\end{tabular}
\end{sc}
\end{small}
\end{center}
\vskip -0.1in
\end{table}

\section{Theoretical Motivation: Risk-Sensitive Control and Epistemic Uncertainty}
\label{app:theoretical_motivation}

We provide a rigorous theoretical justification for the ODAR objective function (Eq. 3). Our formulation is grounded in \textbf{Risk-Sensitive Control Theory}, which naturally yields a mean-variance objective. We then utilize the \textbf{Law of Total Variance} to interpret the variance term as a proxy for epistemic uncertainty.

\subsection{Derivation via Risk-Sensitive Control}
Standard language modeling minimizes the expected surprisal (negative log-likelihood). However, in high-stakes reasoning, we desire an agent that is not only accurate on average but also \textit{risk-averse} to uncertainty.

Let $L(y) = -\log p(y|x)$ denote the loss (surprisal) of a generated sequence $y$. In Risk-Sensitive Control, the objective is to minimize the \textbf{Entropic Risk Measure} (or Exponential Utility), defined as:
\begin{equation}
    \mathcal{J}_\gamma(x) = \frac{1}{\gamma} \log \mathbb{E}_{y \sim p(\cdot|x)} \left[ \exp\left( \gamma L(y) \right) \right]
\end{equation}
where $\gamma > 0$ is the risk-sensitivity parameter. This objective heavily penalizes tails of the loss distribution (i.e., high-surprisal "hallucinations").

Since computing the expectation over the full output space is intractable, we approximate $\mathcal{J}_\gamma$ using a \textbf{second-order Taylor expansion} of the cumulant generating function around the mean $\mu_L = \mathbb{E}[L(y)]$:
\begin{equation}
    \mathcal{J}_\gamma(x) \approx \underbrace{\mathbb{E}[L(y)]}_{\text{Expected Performance}} + \frac{\gamma}{2} \underbrace{\text{Var}[L(y)]}_{\text{Risk Penalty}} + \mathcal{O}(\gamma^2)
    \label{eq:risk_sensitive_expansion}
\end{equation}

\textbf{Conclusion:} The mean-variance form in our objective (Eq. 3) is not a heuristic addition; it is the second-order approximation of the Risk-Sensitive objective. Minimizing $\text{Var}[L(y)]$ is mathematically equivalent to maximizing robustness against worst-case reasoning failures under a risk-averse utility.

\subsection{Interpretation via Variance Decomposition}
Having established \textit{why} we minimize variance (via risk sensitivity), we now analyze \textit{what} this variance represents physically. We employ the \textbf{Law of Total Variance} to decompose the total output volatility based on the latent reasoning state $s$.

Let $L$ be the random variable representing surprisal. We decompose its variance:
\begin{equation}
    \text{Var}[L] = \underbrace{\mathbb{E}_s [ \text{Var}(L | s) ]}_{\text{Aleatoric (Lexical)}} + \underbrace{\text{Var}_s ( \mathbb{E}[L | s] )}_{\text{Epistemic (Reasoning)}}
\end{equation}

Here, conditioning on $s$ implies examining the distribution of $y$ generated by a fixed latent reasoning path.

\textbf{Aleatoric Term:} The first term, $\mathbb{E}_s [ \text{Var}(L | s) ]$, captures variability due to lexical choices (synonyms, phrasing) given a fixed, correct logic. This noise is inherent to language generation.

\textbf{Epistemic Term:} The second term, $\text{Var}_s ( \mathbb{E}[L | s] )$, captures variability in the expected surprisal caused by uncertainty over the reasoning path $s$ itself.

\textbf{Assumption 1 (Epistemic Dominance).} We posit that for rigorous reasoning tasks, the variance induced by switching between contradictory reasoning paths (epistemic) dominates the variance induced by lexical phrasing (aleatoric). That is: $\text{Var}_s ( \mathbb{E}[L | s] ) \gg \mathbb{E}_s [ \text{Var}(L | s) ]$.

Under this assumption, the total variance $\text{Var}[L(y)]$ serves as a tight lower-bound proxy for epistemic uncertainty. Thus, the Risk-Sensitive objective effectively penalizes reasoning ambiguity.

\subsection{ Practical Estimation via Ergodicity}
The derivation above relies on the ensemble variance $\text{Var}[L(y)]$ over the distribution of all possible generations. At inference time, we only have access to a single generated sequence $y_t$.

To make this computable, we introduce a \textbf{Practical Estimator} based on the \textit{Weak Ergodicity Hypothesis}. We assume that for a long, coherent reasoning chain, the temporal variance of token-level surprisal approximates the ensemble variance of the generation process:
\begin{equation}
    \text{Var}_{y \sim p(\cdot|x)}[L(y)] \approx \widehat{\text{Var}}_t [ -\log p(y_t | y_{<t}, x) ]
\end{equation}

\textbf{Robust Implementation Note:} To satisfy the stationarity requirement implied by ergodicity, we exclude highly structured tokens (e.g., formatting templates, repeated headers) from the variance calculation, computing $\widehat{\text{Var}}_t$ only over the substantive reasoning steps. This prevents format-induced volatility from confounding the epistemic signal.

\subsection{Summary of the Theoretical Argument}
Our framework is rigorously grounded in the following logic flow. \textbf{1. Origin:} We adopt a \textbf{Risk-Sensitive} control objective (Exponential Utility). \textbf{2. Derivation:} A \textbf{Taylor expansion} yields a tractable Mean-Variance objective. \textbf{3. Interpretation:} Via \textbf{Variance Decomposition}, minimizing variance targets Epistemic Uncertainty (hallucinations). \textbf{4. Implementation:} We operationalize this via a \textbf{temporal variance estimator} over the reasoning chain.
\section{Algorithm Specifications}
\label{app:algorithms}

This appendix provides the complete algorithmic specifications for ODAR's core components. We first present the FEP-based multi-agent fusion logic used for candidate selection, followed by the adaptive routing procedure that governs the overall inference flow. 

\begin{algorithm}[H]
\caption{FEP-Based Multi-Agent Fusion (Character-Level \& Z-Score)}
\label{alg:fep_fusion_app}
\small
\begin{algorithmic}[1]
\REQUIRE Candidate set $\mathcal{Y} = \{y_1, \ldots, y_K\}$, query $x$, penalty weight $\lambda = 0.1$ 
\REQUIRE Model statistics $\mu_m, \sigma_m$ for each agent $m$ (for Z-score alignment)
\ENSURE Selected answer $\hat{y}$

\STATE // \textbf{Step 1: Compute Character-Normalized Free Energy}
\FOR{$i = 1$ to $K$}
    \STATE // Calculate Energy Density per Character (Eq. 10) 
    \STATE $\ell_i \gets -\frac{1}{|y_i|_{\text{char}}} \sum_{t=1}^{|y_i|_{\text{tok}}} \log p(y_{i,t} \mid y_{i,<t}, x)$ 
    
    \STATE // Calculate Uncertainty Penalty (Varentropy) 
    \STATE $\sigma^2_i \gets \text{Var}[\log p(y_{i,t} \mid y_{i,<t}, x)]$ 
    
    \STATE // Raw Free Energy (Character-normalized) 
    \STATE $\mathcal{F}_{\text{raw}}(y_i|x) \gets \ell_i + \lambda \cdot \sigma^2_i$
    
    \STATE // Model-Specific Z-Score Alignment (Eq. 11) 
    \STATE $\mathcal{F}(y_i|x) \gets \frac{\mathcal{F}_{\text{raw}}(y_i|x) - \mu_m}{\sigma_m}$
\ENDFOR

\STATE // \textbf{Step 2: Select candidate with minimum standardized energy}
\STATE $\hat{y} \gets \arg\min_{y_i \in \mathcal{Y}} \mathcal{F}(y_i|x)$ 
\STATE \textbf{Return} $\hat{y}$
\end{algorithmic}
\end{algorithm}

\vspace{1em} 

\begin{algorithm}[H]
\caption{Active Inference for Adaptive Routing}
\label{alg:active_inference_app}
\small
\begin{algorithmic}[1]
\REQUIRE Query $x$, Difficulty Estimator $\text{DE}(\cdot)$, thresholds $\tau_1 = 0.3$, $\tau_2 = 0.7$ 
\ENSURE Selected policy $\pi^*$ and final answer $\hat{y}$

\STATE // \textbf{Step 1: Estimate difficulty (approximate EFE)}
\STATE $d \gets \text{DE}(x)$ 

\STATE // \textbf{Step 2: Select policy based on difficulty} 
\IF{$d < \tau_1$} 
    \STATE $\pi^* \gets \pi_{\text{S}}$ \hfill // Simple Path: low epistemic value 
\ELSIF{$d < \tau_2$}
    \STATE $\pi^* \gets \pi_{\text{M}}$ \hfill // Medium Path: moderate epistemic value 
\ELSE
    \STATE $\pi^* \gets \pi_{\text{H}}$ \hfill // Hard Path: high epistemic value 
\ENDIF

\STATE // \textbf{Step 3: Execute selected policy}
\IF{$\pi^* = \pi_{\text{S}}$}
    \STATE $\hat{y} \gets m_\theta(x)$ \hfill // Fast Agent only 
\ELSIF{$\pi^* = \pi_{\text{M}}$}
    \STATE $y_\theta \gets m_\theta(x)$, $y_\gamma \gets m_\gamma(x, y_\theta)$ \hfill // Slow Agent verifies 
    \STATE $\hat{y} \gets \text{FEP-Fusion}(\{y_\theta, y_\gamma\})$ \hfill // See Algorithm~\ref{alg:fep_fusion_app} 
\ELSE
    \STATE $y_\theta \gets m_\theta(x)$ 
    \STATE $\mathcal{Y}_\gamma \gets \{m_\gamma(x)_i : i = 1, \ldots, n\}$ \hfill // Best-of-N ($n=5$) 
    \STATE $\hat{y} \gets \text{FEP-Fusion}(\{y_\theta\} \cup \mathcal{Y}_\gamma)$ \hfill // See Algorithm~\ref{alg:fep_fusion_app} 
\ENDIF
\STATE \textbf{return} $\pi^*, \hat{y}$
\end{algorithmic}
\end{algorithm}

\section{Difficulty Estimator Training Details}
\label{app:de_training}

\subsection{Data Construction for Empirical EFE}
Since the definition of the Empirical EFE target $d^*$ (see Eq.~\ref{eq:empirical_efe} in Section~\ref{sec:de}) requires ground-truth labels for both human cognition time and model accuracy, we constructed a specialized calibration dataset $\mathcal{D}_{cal} = \{(x_i, d_i^*)\}_{i=1}^{N}$ with $N=2,000$ samples, stratified across our benchmark suite.

\paragraph{Human Time Proxy ($T_{\text{human}}$).} 
To approximate the epistemic uncertainty term, we sourced average solution times from metadata provided in the ARC-Challenge, MATH, and GPQA datasets. For datasets without explicit human timing logs (e.g., GSM8K), we utilized the normalized token length of the ground-truth Chain-of-Thought (CoT) reasoning as a proxy for temporal cognitive load. We assume that the length of a valid reasoning chain correlates positively with the intrinsic entropy of the problem state.

\paragraph{Fast Agent Accuracy ($Acc_{\text{fast}}$).} 
To measure the pragmatic risk term, we executed the Fast Agent ($A_\theta$, GPT-5.1 with $T=0.2$) on all training samples to obtain binary correctness labels $\mathbb{I}(y_{\theta} == y_{GT})$. A failure ($Acc=0$) signals a high-risk state necessitating the Slow Agent's intervention.

\subsection{Optimization Configuration}
The estimator parameters $W \in \mathbb{R}^{1 \times 24}$ and $b \in \mathbb{R}$ are optimized to minimize the Mean Squared Error (MSE) between the predicted amortized energy $d(x)$ and the empirical target $d^*$. We incorporated $L_2$ regularization to prevent overfitting to specific dataset keywords:
\begin{equation}
\mathcal{L}_{\text{DE}} = \frac{1}{N} \sum_{i=1}^N (d(x_i) - d_i^*)^2 + \lambda_{\text{reg}} \|W\|_2^2
\label{eq:de_loss_app}
\end{equation}
Training was performed using the Adam optimizer with a learning rate of $10^{-3}$, a batch size of 32, and $\lambda_{\text{reg}}=0.01$ for 50 epochs. The feature extraction logic ($\phi(x)$) is detailed in Appendix~\ref{app:feature_extraction}.

\section{Neuro-Computational Motivation}
\label{app:neuro_motivation}

\subsection{Biological Inspiration: Threshold-Gated Conditional Computation}
\label{app:neuro_background}

Artificial neural networks and biological brains operate on fundamentally different substrates—continuous oscillatory dynamics across billions of neurons versus discrete routing between two software agents. ODAR does not attempt to simulate biological neural mechanisms. Instead, we draw a \textbf{functional analogy} to the brain's Theta-Gamma Phase-Amplitude Coupling (TG-PAC) at the level of \textit{control logic}, abstracting a single design principle: \textbf{threshold-gated conditional computation}.

Recent single-neuron recordings in humans  suggest that TG-PAC implements a form of conditional computation in biological circuits:
\begin{itemize}
    \item \textbf{Theta Phase as Control Signal:} The slow theta rhythm (4-8 Hz) acts as a temporal gate, organizing neural firing into discrete windows of processing. It reflects cognitive control demand.
    \item \textbf{Gamma Amplitude as Gated Resource:} High-frequency gamma bursts (30-100 Hz) represent intensive local processing. Crucially, these metabolically expensive bursts are not continuous; they are selectively triggered only when the theta phase indicates a high-demand condition.
\end{itemize}

\textbf{What ODAR Abstracts:} The core principle that expensive computation should be \textit{conditionally gated} by a control signal reflecting task demand. Formally, both systems implement a control law of the form:
\begin{equation}
\text{Resource}(t) = \text{Base} + \mathbb{I}(\text{ControlSignal}(t) > \text{Threshold}) \cdot \text{ExpensiveCompute}
\end{equation}
where $\mathbb{I}(\cdot)$ is the indicator function. In TG-PAC, the control signal is oscillatory theta phase; in ODAR, it is the learned difficulty score $\mathcal{D}(x)$.

\textbf{What ODAR Does Not Claim:}
\begin{itemize}
    \item \textit{Temporal dynamics:} ODAR's routing is aperiodic and input-driven, not oscillatory.
    \item \textit{Neural implementation:} ODAR uses two discrete LLM agents, not billions of coupled neurons.
    \item \textit{Mechanistic equivalence:} The analogy is at the level of control flow, not biophysics.
\end{itemize}

This "gating-for-efficiency" principle motivates ODAR's architecture: the \textit{Difficulty Estimator} acts as the control signal, gating invocation of the computationally expensive \textit{Slow Agent} only when the estimated difficulty exceeds threshold $\tau$. A formal analysis of this functional correspondence is provided in Appendix~\ref{app:neuro_abstraction}.

\subsection{Algorithmic Abstraction: From Oscillatory Gating to Discrete Routing}
\label{app:neuro_abstraction}

To rigorously justify this bio-inspiration without claiming biological realism, we define ODAR as a \textbf{Discrete-Event Abstraction} of the TG-PAC dynamics. We show that both systems implement the same \textbf{Threshold-Gated Control Law}.

\textbf{1. The Biological Reference System (Continuous)}
The governing control law for neural resource recruitment can be modeled as :
\begin{equation}
A_\gamma(t) = A_{base} + \underbrace{\mathcal{H}(u_{bio}(t) - \kappa)}_{\text{Gating Function}} \cdot A_{burst}
\label{eq:bio_control}
\end{equation}
where $A_\gamma$ is the local activity amplitude, $\mathcal{H}$ is the Heaviside step function, $u_{bio}(t) = \sin(\phi_\theta(t))$ is the oscillatory control signal, and $\kappa$ is the activation threshold.

\textbf{2. The ODAR System (Discrete)}
ODAR operates on a sequence of discrete queries $x_i$. The computational cost $C(x_i)$ is determined by the router's decision:
\begin{equation}
C(x_i) = C_{base} + \underbrace{\mathbb{I}(\mathcal{D}(x_i) - \tau)}_{\text{Routing Function}} \cdot C_{delta}
\label{eq:odar_control}
\end{equation}
where $\mathbb{I}$ is the indicator function, $\mathcal{D}(x_i)$ is the difficulty score (control signal), and $\tau$ is the learned threshold.

\textbf{3. Functional Homomorphism}
We establish a structure-preserving map between the control variables:
\begin{itemize}
    \item \textbf{Control Signal Mapping ($u(t) \mapsto \mathcal{D}(x_i)$):} In the brain, theta phase dictates the "window of excitability". In ODAR, the difficulty score dictates the "window of necessity". Both serve as the excitability signal $u$.
    \item \textbf{Actuation Mapping ($A_{burst} \mapsto C_{delta}$):} Both represent the recruitment of metabolically expensive resources (spiking bursts vs. API tokens).
\end{itemize}

\textbf{Proposition (Objective Equivalence):} 
If we view the stream of queries $x_1, \dots, x_N$ as sampling a continuous cognitive load signal, ODAR's routing objective is the \textbf{Riemann Sum approximation} of the biological energy-efficiency functional:
\begin{equation}
J_{bio} \approx \sum_{i} [Acc(x_i) - \lambda \cdot \mathbb{I}(u(x_i) - \tau)]
\end{equation}
This demonstrates that ODAR is algorithmically equivalent to a discretized solution of the biological resource allocation problem, replacing temporal periodicity (required for biological sync) with semantic gating (required for software efficiency).

\begin{table}[ht]
\caption{Functional Correspondence between TG-PAC and ODAR.}
\label{tab:neuro_mapping}
\begin{center}
\begin{small}
\begin{sc}
\begin{tabular}{lll}
\toprule
Biological Component & ODAR Abstraction & Computational Role \\
\midrule
Theta Rhythm (4-8Hz) & Difficulty Estimator & \textbf{Gating Signal:} Assesses demand/load \\
Gamma Bursts (30-100Hz) & Slow Agent Execution & \textbf{Intensive Compute:} Expensive processing \\
Phase-Amplitude Coupling & Dynamic Routing ($\tau$) & \textbf{Conditional Logic:} If High Load $\to$ Activate \\
Metabolic Constraints & Token Cost Budget & \textbf{Constraint:} Maximize accuracy per unit cost \\
\bottomrule
\end{tabular}
\end{sc}
\end{small}
\end{center}
\end{table}
\section{Extended Comparison with Concurrent Routing Frameworks}
\label{app:qualitative_comparison}

To situate ODAR within the rapidly evolving landscape of adaptive inference, we provide a qualitative comparison with recent concurrent works in Table~\ref{tab:routing_comparison}.

\begin{table*}[ht]
\centering
\caption{\textbf{Qualitative Comparison of Adaptive Routing Frameworks.} ODAR distinguishes itself by integrating principled difficulty estimation (Active Inference) with a training-free fusion mechanism, avoiding the high training overhead of RL-based methods while targeting reasoning depth specifically.}
\label{tab:routing_comparison}
\resizebox{\textwidth}{!}{%
\begin{tabular}{lcccccc}
\toprule
\textbf{Framework} & \textbf{Core Mechanism} & \textbf{Router Training} & \textbf{Inference Overhead} & \textbf{Primary Goal} & \textbf{Reasoning Focus} & \textbf{Bio-Inspired} \\
\midrule
\textbf{HierRouter} \cite{55} & Reinforcement Learning (PPO) & High (Policy Training) & Medium & General Efficiency & Partial & No \\
\textbf{RTR} \cite{53} & Supervised Regression & Medium (Scalar Predictor) & Low & Strategy Selection & Yes & No \\
\textbf{MoMA} \cite{54} & Intent Classification & Medium (Capability Profiling) & Low & General Orchestration & No & No \\
\textbf{CP-Router} \cite{56} & Conformal Prediction & Low (Calibration Only) & Medium (Sampling) & Uncertainty/Cost & Yes & No \\
\textbf{InferenceDynamics} \cite{57} & Capability Knowledge Graph & Medium (Profiling) & Low & Knowledge Routing & No & No \\
\midrule
\rowcolor{gray!10} \textbf{ODAR (Ours)} & \textbf{Active Inference \& FEP} & \textbf{Low (Lightweight DE)} & \textbf{Low} & \textbf{Test-Time Scaling} & \textbf{Strong (System 2)} & \textbf{Yes (TG-PAC)} \\
\bottomrule
\end{tabular}%
}
\end{table*}

\paragraph{Key Differentiators.}
ODAR differentiates itself from existing frameworks through three key dimensions. First, in terms of \textbf{training efficiency}, unlike HierRouter which necessitates expensive Reinforcement Learning policy optimization, ODAR employs a lightweight Difficulty Estimator trained on readily available metadata, coupled with an entirely training-free fusion mechanism. Second, regarding \textbf{theoretical grounding}, while approaches like CP-Router rely on statistical uncertainty via conformal prediction, ODAR uniquely operationalizes the Free Energy Principle. By using Varentropy as a proxy for epistemic uncertainty, it specifically targets and penalizes hallucinated reasoning paths rather than just generic confidence intervals. Finally, ODAR prioritizes \textbf{System 2 specialization}. Whereas frameworks like MoMA and InferenceDynamics focus on domain-based routing (e.g., dispatching Math vs. History queries), ODAR optimizes routing across \textit{reasoning depths} (Fast heuristics vs. Slow deliberation), directly addressing the "overthinking" phenomenon in complex logical tasks.

\section{Theory-to-Implementation Mapping}
\label{app:theory_to_impl}

\begin{figure}[ht]
\centering
\includegraphics[width=\linewidth]{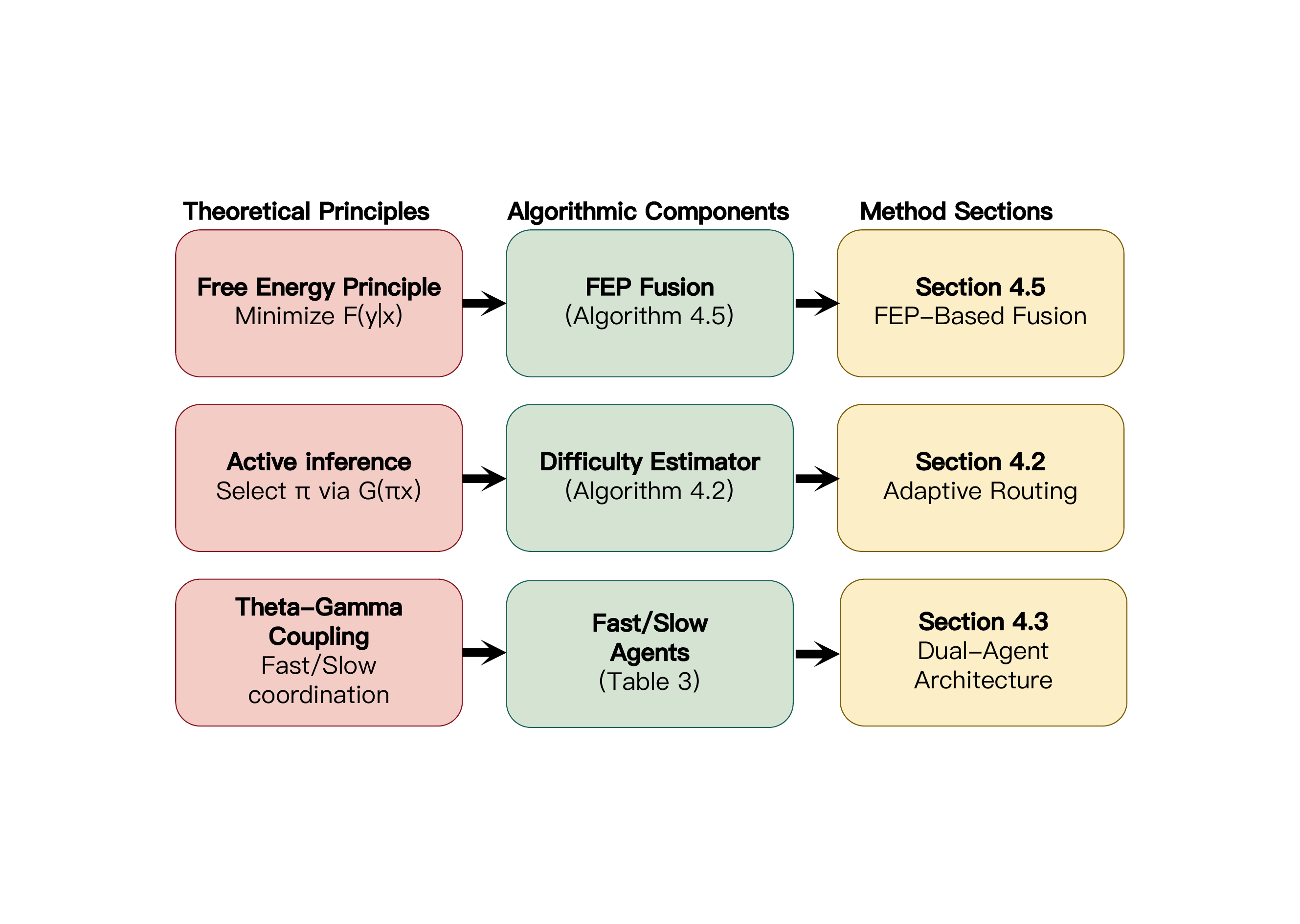}
\caption{Mapping from theoretical principles to computational implementation. Each theoretical framework contributes a distinct component: FEP enables principled fusion, Active Inference guides adaptive routing, and theta-gamma coupling motivates the dual-agent architecture.}
\label{fig:theory_to_method_app}
\end{figure}

Figure~\ref{fig:theory_to_method_app} illustrates the correspondence between theoretical principles and ODAR's algorithmic components. The Free Energy Principle motivates the FEP-based fusion mechanism (Section~3.5), which selects answers by minimizing variational free energy rather than relying on heuristic voting. Active Inference theory guides the design of adaptive routing (Section~3.2), where the Difficulty Estimator approximates Expected Free Energy to select among reasoning policies. The theta-gamma coupling literature justifies the Fast-Slow agent dichotomy (Section~3.3), providing a biologically-grounded inductive bias for heterogeneous agent specialization.

\begin{figure}[ht]
\centering
\includegraphics[width=\linewidth]{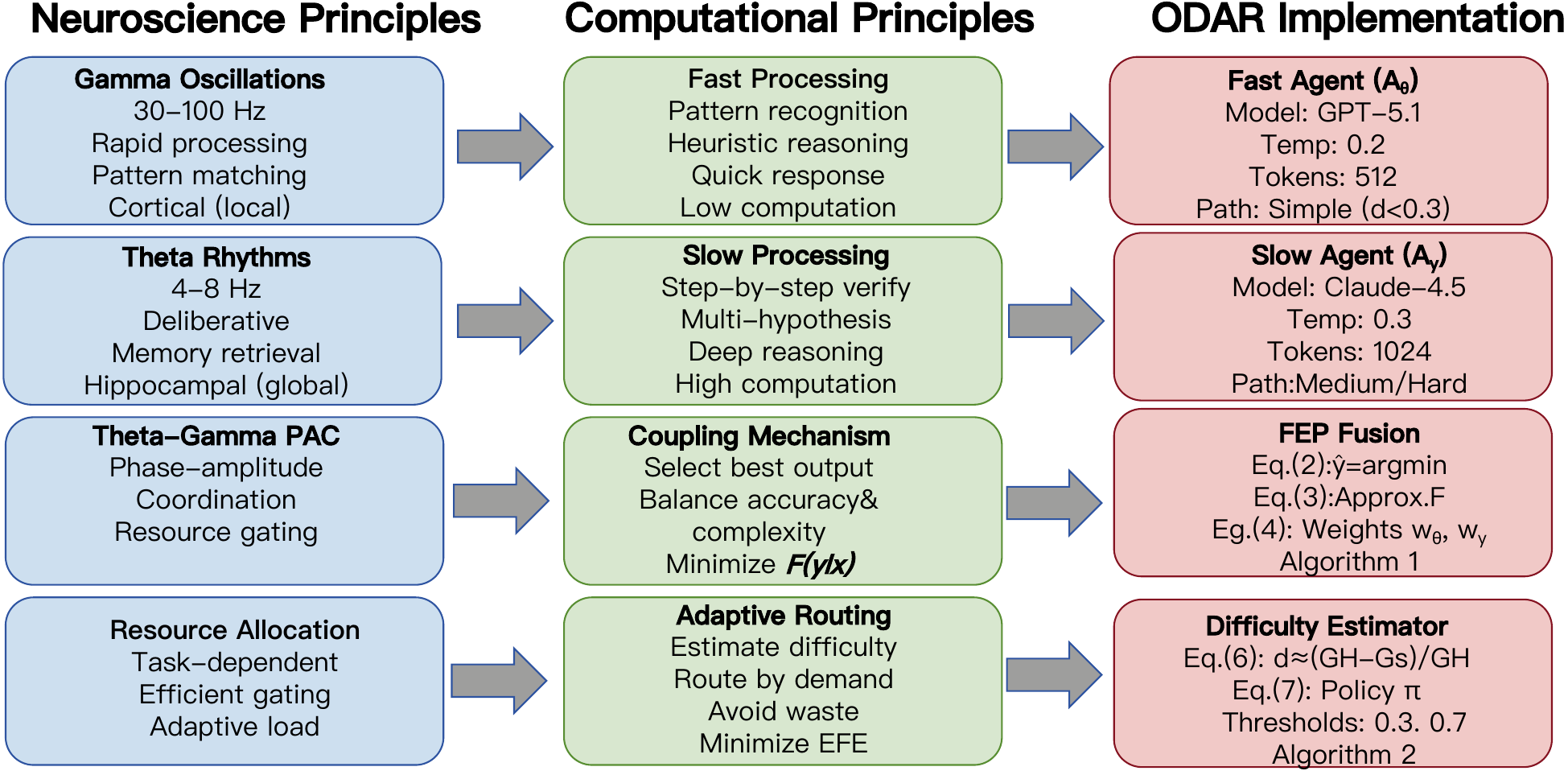}
\caption{Theoretical framework overview. Left panel: neuroscientific principles including gamma oscillations, theta rhythms, and the Free Energy Principle. Middle panel: corresponding computational principles of fast pattern matching, slow verification, and uncertainty-driven resource allocation. Right panel: ODAR implementation comprising the Fast Agent, Slow Agent, and Difficulty Estimator.}
\label{fig:theory_overview_app}
\end{figure}
\section{Feature Extraction for Difficulty Estimation}
\label{app:feature_extraction}

To train the Difficulty Estimator, we extract a comprehensive feature vector $\phi(x) \in \mathbb{R}^{24}$ encoding lexical, structural, and semantic properties of query $x$. This section provides complete specifications for reproducibility.

\subsection{Feature Specification}

Table~\ref{tab:feature_complete} presents the complete feature specification with definitions and importance scores derived from permutation importance analysis on the validation set.

\begin{table}[ht]
\centering
\caption{Complete feature specification for the Difficulty Estimator. Importance scores represent mean decrease in Pearson $r$ when feature is permuted, averaged over 10 runs.}
\label{tab:feature_complete}
\small
\begin{tabular}{@{}clp{6.5cm}c@{}}
\toprule
\textbf{ID} & \textbf{Feature} & \textbf{Definition} & \textbf{Imp.} \\
\midrule
\multicolumn{4}{l}{\textit{Lexical Features ($k_1 = 8$)}} \\
\midrule
$f_1$ & Character length & Total number of characters in query & 0.012 \\
$f_2$ & Word count & Number of whitespace-separated tokens & 0.018 \\
$f_3$ & Token count & BPE tokens via \texttt{tiktoken} (cl100k\_base) & 0.024 \\
$f_4$ & Avg. word length & Mean character count per word & 0.008 \\
$f_5$ & Max word length & Maximum character count across words & 0.006 \\
$f_6$ & Flesch-Kincaid & Readability grade level, clamped to $[0, 20]$ & 0.015 \\
$f_7$ & Numeric indicator & Binary: contains digits & 0.031 \\
$f_8$ & Math symbol indicator & Binary: contains symbols from $\mathcal{M}$ & 0.042 \\
\midrule
\multicolumn{4}{l}{\textit{Structural Features ($k_2 = 6$)}} \\
\midrule
$f_9$ & Sentence count & Via spaCy sentence segmentation & 0.019 \\
$f_{10}$ & Clause count & SBAR/SBARQ nodes in constituency parse & 0.027 \\
$f_{11}$ & Question marks & Count of ``?'' characters & 0.009 \\
$f_{12}$ & Image indicator & Binary: multimodal input present & 0.038 \\
$f_{13}$ & Parse tree depth & Maximum depth in dependency parse & 0.033 \\
$f_{14}$ & Logical connectives & Count of terms from $\mathcal{L}$ & 0.022 \\
\midrule
\multicolumn{4}{l}{\textit{Semantic Features ($k_3 = 10$)}} \\
\midrule
$f_{15}$ & TF-IDF (math) & Aggregated TF-IDF for math vocabulary & 0.048 \\
$f_{16}$ & TF-IDF (code) & Aggregated TF-IDF for coding vocabulary & 0.041 \\
$f_{17}$ & TF-IDF (logic) & Aggregated TF-IDF for logic vocabulary & 0.035 \\
$f_{18}$ & Difficulty density & Fraction of words matching $\mathcal{H}$ & 0.052 \\
$f_{19}$ & Type: multiple-choice & Binary: enumerated answer options present & 0.029 \\
$f_{20}$ & Type: open-ended & Binary: open response format & 0.025 \\
$f_{21}$ & Type: proof & Binary: proof/derivation requested & 0.044 \\
$f_{22}$ & Domain: math & Softmax probability from 8-way classifier & 0.056 \\
$f_{23}$ & Domain: code & Softmax probability from 8-way classifier & 0.047 \\
$f_{24}$ & Domain: reasoning & Softmax probability from 8-way classifier & 0.039 \\
\bottomrule
\end{tabular}
\end{table}

\subsection{Lexical Features}

\paragraph{Length Statistics ($f_1$--$f_5$).} 
We compute character count, word count, and token count using the tiktoken library with the cl100k encoding to ensure consistency with GPT-5.1 tokenization. Average and maximum word lengths capture lexical complexity.

\paragraph{Readability ($f_6$).} 
The Flesch-Kincaid grade level is computed via the textstat library (v0.7.3). For queries containing mathematical notation, we replace LaTeX expressions with placeholder tokens before syllable counting. Scores are clamped to $[0, 20]$.

\paragraph{Mathematical Content ($f_7$, $f_8$).} 
We define the mathematical symbol set as:
\begin{equation}
\mathcal{M} = \{+, -, \times, \div, =, \neq, <, >, \leq, \geq, \sum, \prod, \int, \partial, \nabla, \infty\} \cup \{\text{LaTeX commands}\}
\end{equation}
Feature $f_8$ indicates whether any symbol from $\mathcal{M}$ appears in the query.

\subsection{Structural Features}

\paragraph{Syntactic Complexity ($f_9$, $f_{10}$, $f_{13}$).} 
We employ spaCy  for sentence segmentation and dependency parsing. Clause counting uses the Berkeley Neural Parser to identify subordinate clause nodes (SBAR, SBARQ) in constituency parses. For mathematical content, we apply preprocessing that replaces LaTeX blocks with placeholder tokens before parsing, improving parser accuracy from 67\% to 94\% on math-heavy queries.

\paragraph{Logical Structure ($f_{14}$).} 
We define the logical connective set as:
\begin{equation}
\mathcal{L} = \{\textit{if, then, therefore, hence, thus, because, since, implies, iff, for all, there exists, such that}\}
\end{equation}
Feature $f_{14}$ counts case-insensitive occurrences of terms from $\mathcal{L}$.

\subsection{Semantic Features}

\paragraph{Domain-Specific TF-IDF ($f_{15}$--$f_{17}$).} 
We construct domain vocabularies from 10,000 annotated queries in our calibration set. For each domain $d \in \{\text{math}, \text{code}, \text{logic}\}$, we extract the top-100 discriminative terms by TF-IDF, excluding stopwords and high-frequency terms ($>$50\% of queries). Representative terms include:

\begin{table}[ht]
\centering
\caption{Representative vocabulary by domain used in lexical feature extraction.}
\label{tab:vocab_domain}
\setlength{\tabcolsep}{0pt} 
\begin{tabular*}{\linewidth}{@{\extracolsep{\fill}}ll@{}}
\toprule
\textbf{Domain} & \textbf{Representative Vocabulary ($|V_d|$)} \\
\midrule
Mathematics (87) & \textit{theorem, derivative, integral, matrix, eigenvalue, convergence, polynomial} \\
Coding (92) & \textit{algorithm, function, complexity, implement, recursive, runtime, optimize} \\
Logic (73) & \textit{valid, satisfiable, contradiction, premise, inference, deduction, entailment} \\
\bottomrule
\end{tabular*}
\end{table}
\noindent The query-level score is computed as $f_d(x) = \sum_{t \in x \cap V_d} \text{tfidf}(t)$.

\paragraph{Difficulty Markers ($f_{18}$).} 
We curate a set $\mathcal{H}$ of 47 high-difficulty markers based on manual analysis of 500 challenging problems. The set includes cognitive demand indicators such as \textit{prove}, \textit{derive}, \textit{generalize}, \textit{optimize}, \textit{compare and contrast}, \textit{analyze}, and \textit{synthesize}. Difficulty density is computed as:
\begin{equation}
f_{18} = \frac{|\{h \in \mathcal{H} : h \subseteq x\}|}{|x|_{\text{words}}}
\end{equation}

\paragraph{Question Type ($f_{19}$--$f_{21}$).} 
We classify question format using pattern matching: multiple-choice queries exhibit enumerated options (e.g., ``(A)'', ``(B)''); open-ended queries begin with interrogative words without option structure; proof requests contain explicit derivation keywords.

\paragraph{Domain Classification ($f_{22}$--$f_{24}$).} 
We train a DistilBERT-based 8-way classifier (66M parameters) on 50,000 labeled queries, achieving 89.2\% accuracy. The eight classes are: Mathematics, Coding, Logic, Commonsense, Knowledge, Multimodal, Language, and Other. Features $f_{22}$--$f_{24}$ extract softmax probabilities for the three most predictive domains.

\subsection{Feature Importance Analysis}

Table~\ref{tab:feature_importance} presents the top-10 features ranked by permutation importance.

\begin{table}[ht]
\centering
\caption{Top-10 features by permutation importance. Semantic features contribute 56\% of total importance.}
\label{tab:feature_importance}
\setlength{\tabcolsep}{0pt} 
\begin{tabular*}{\linewidth}{@{\extracolsep{\fill}}clcc@{}}
\toprule
\textbf{Rank} & \textbf{Feature} & \textbf{Category} & \textbf{Importance} \\
\midrule
1 & Domain: math ($f_{22}$) & Semantic & 0.056 \\
2 & Difficulty density ($f_{18}$) & Semantic & 0.052 \\
3 & TF-IDF math ($f_{15}$) & Semantic & 0.048 \\
4 & Domain: code ($f_{23}$) & Semantic & 0.047 \\
5 & Type: proof ($f_{21}$) & Semantic & 0.044 \\
6 & Math symbol ($f_8$) & Lexical & 0.042 \\
7 & TF-IDF code ($f_{16}$) & Semantic & 0.041 \\
8 & Domain: reasoning ($f_{24}$) & Semantic & 0.039 \\
9 & Image indicator ($f_{12}$) & Structural & 0.038 \\
10 & TF-IDF logic ($f_{17}$) & Semantic & 0.035 \\
\bottomrule
\end{tabular*}
\end{table}

The analysis reveals that semantic features dominate difficulty prediction, with the domain classifier outputs ($f_{22}$--$f_{24}$) and difficulty density ($f_{18}$) providing the strongest signal. Lexical and structural features serve as complementary indicators that improve robustness on out-of-distribution queries.

\subsection{Ablation Study}

Table~\ref{tab:feature_ablation} presents results when removing each feature category.

\begin{table}[ht]
\centering
\caption{Feature category ablation. Removing semantic features causes the largest performance degradation.}
\label{tab:feature_ablation}
\setlength{\tabcolsep}{0pt} 
\begin{tabular*}{\linewidth}{@{\extracolsep{\fill}}lccc@{}}
\toprule
\textbf{Configuration} & \textbf{Pearson $r$} & \textbf{Routing Acc.} & \textbf{Task Acc.} \\
\midrule
Full model (24 features) & 0.79 & 87.3\% & 83.5\% \\
Without lexical ($k_1=8$) & 0.76 & 85.1\% & 82.8\% \\
Without structural ($k_2=6$) & 0.74 & 83.8\% & 82.1\% \\
Without semantic ($k_3=10$) & 0.61 & 74.2\% & 79.3\% \\
\midrule
Random baseline & 0.00 & 33.3\% & 75.4\% \\
\bottomrule
\end{tabular*}
\end{table}

Removing semantic features reduces routing accuracy by 13.1 percentage points and end-to-end task accuracy by 4.2 points, confirming their critical role. The three-tier design provides necessary redundancy: lexical features capture surface complexity, structural features encode syntactic depth, and semantic features identify domain-specific difficulty patterns.
\subsection{Extended Ablation Studies and Systematic Analysis}
\label{app:extended_ablations}

To further disentangle the contributions of the learned Difficulty Estimator (DE) from fixed orchestration heuristics and compute-time scaling, we present a series of rigorous ablation studies.

\subsubsection{Formal Definition and Impact of the Rule-based Orchestration Layer}
\label{app:rule_definition}

As described in Section~3, ODAR includes a lightweight rule-based dispatch/orchestration layer (ER/MR/SS) implemented with fixed heuristics. For clarity, we refer to \textbf{Expert Router (ER)} and \textbf{Model Router (MR)} as the \emph{dispatch heuristics}, which identify modality/domain cues to assign expert types ($e$) and select base-model identifiers ($m$). The \textbf{Strategy Selector (SS)} acts as the \emph{difficulty-gated path selector}, mapping the DE-produced score $d$ and features to a final reasoning path and call-budget $c \in \{1,2,6\}$.

\paragraph{Priority and Conflict Resolution.}
Within the orchestration heuristics, conflicts are resolved via a fixed priority hierarchy as shown in Figure~1: \textbf{Modality $\rightarrow$ Keywords $\rightarrow$ Question Type $\rightarrow$ Fallback}. When multiple cues within the same level fire, ties are broken by keyword-count voting and a fixed domain priority list. This ensures a deterministic $(e,m)$ configuration for each query. The difficulty score $d$ remains a product of the learned DE, and path-gating depends on thresholds $(\tau_1, \tau_2)$.

\subsection{De-confounding Ablation I: The "Heuristics-Off" Baseline}
To assess whether ODAR's gains stem primarily from expert/model selection heuristics, we evaluate a version of ODAR where \textbf{ER/MR (heuristic dispatch) are disabled}, while keeping the same \textbf{DE + threshold gating} (SS) used in the main system. In this setup, we fix the expert configuration to "general" and bypass MR's model switching, maintaining the default dual-agent configuration (\textbf{Fast=GPT-5.1, Slow=Claude-4.5}) across all tasks.

As shown in Table~\ref{tab:ablation_deconfounding}, disabling ER/MR heuristics results in a minor performance drop ($-1.2\%$ macro average). This confirms that while the orchestration layer optimizes configuration-specific efficiency, the \textbf{learned DE-based routing is the primary driver} of the system's performance.

\subsection{De-confounding Ablation II: Proportion-matched Randomized Routing}
To isolate the quality of the decision logic, we implement a \textbf{Proportion-matched Randomized Routing} baseline. 
\begin{itemize}
    \item \textbf{Setup:} We maintain the same path proportions (e.g., 41\% Simple, 35\% Medium, 24\% Hard) as the full ODAR system, \textbf{matched to the global routing proportions measured over the entire evaluation mixture}. This ensures an \textbf{identical aggregate call-budget} (same $c$ and token limits). However, the $d/\text{SS}$ routing decision is replaced by a random assignment following these probabilities. All other factors—including ER/MR configurations and the Risk-Sensitive Fusion (FEP)—remain identical.
    \item \textbf{Result:} Learned routing outperforms randomized routing by \textbf{+8.3\%} on macro average (Table~\ref{tab:ablation_deconfounding}). This proves that ODAR's advantage stems from the DE's ability to accurately identify \textit{which} specific queries require deliberative reasoning, rather than simply increasing the total number of calls.
\end{itemize}

\begin{table*}[ht]
\centering
\caption{Step-wise Ablation and De-confounding Analysis. All results are averaged over 3 seeds; subsample uncertainty is reported as t-based 95\% CI over $K=5$ disjoint folds. The "Rules Only" baseline uses fixed routing (all queries to Medium path via overriding $d \equiv 0.5$) with ER/MR active, thus bypassing Hard-path Best-of-N expansion.}
\label{tab:ablation_deconfounding}
\setlength{\tabcolsep}{0pt}
\small
\begin{tabular*}{\textwidth}{@{\extracolsep{\fill}}lccccc@{}}
\toprule
\textbf{Configuration} & \textbf{Macro Avg.} & \textbf{MMLU-Pro} & \textbf{MATH} & \textbf{HLE} & \textbf{Subsample CI} \\
\midrule
Base (Single Model) & 77.6\% & 89.5\% & 91.8\% & 42.8\% & $\pm$0.32\% \\
ODAR (Rules Only, No DE Gating) & 79.2\% & 90.1\% & 93.4\% & 44.1\% & $\pm$0.41\% \\
ODAR (ER/MR Off, DE Active) & 88.4\% & 93.1\% & 96.9\% & 53.2\% & $\pm$0.55\% \\
\rowcolor{gray!10} \textbf{Full ODAR (Final)} & \textbf{89.6\%} & \textbf{94.2\%} & \textbf{98.2\%} & \textbf{54.8\%} & \textbf{$\pm$0.45\%} \\
\midrule
Randomized (Matched Call-Budget) & 81.3\% & 91.2\% & 94.1\% & 46.5\% & $\pm$0.85\% \\
\bottomrule
\end{tabular*}
\end{table*}

\subsection{Implementation}

Feature extraction achieves throughput exceeding 1,000 queries per second on a single CPU core, ensuring the Difficulty Estimator does not become a latency bottleneck. Key dependencies include tiktoken (v0.5.1), textstat (v0.7.3), spacy (v3.5.0), and transformers (v4.30.0). Complete extraction code is provided in the supplementary materials.
\section{Error Analysis and Taxonomy}
\label{app:error_analysis}

To better understand ODAR's failure modes, we performed a fine-grained manual analysis on a subset of incorrect predictions.
Because long-horizon reasoning traces are labor-intensive to inspect, we construct an error pool consisting of all \emph{incorrect} model outputs across the 23 evaluation datasets, and then uniformly sample $N{=}100$ error instances from this pool.\footnote{We treat an instance as an error if the final returned answer $\hat{y}$ is judged incorrect by the dataset's official metric (exact match or task-specific evaluator).}
Each sampled instance is annotated by two annotators using a shared guideline; disagreements are resolved through discussion to obtain a final label.\footnote{The taxonomy is designed to separate \emph{system-level} failures introduced by routing, verification, and selection (E1--E4) from \emph{underlying model capability} limitations (E5--E6).}

Table~\ref{tab:error_taxonomy_app} summarizes the resulting taxonomy and its empirical distribution.
In particular, E2 (Expert Routing) corresponds to failures of the \emph{rule-based dispatch layer} (ER/MR) used in our system pipeline: the dispatch assigns an inappropriate expert tag or modality/domain label (e.g., treating a code-generation query as a math-proof query), which then leads to a suboptimal model/prompt/strategy configuration.
E4 (FEP Selection) captures cases where a correct candidate exists in the candidate set (oracle-hit), but the free-energy score ranks an incorrect candidate lower.

\begin{table}[ht]
\centering
\caption{Error Taxonomy and Distribution ($N=100$). Errors are categorized to distinguish between system-level routing failures (E1--E4) and underlying model capability limitations (E5--E6).}
\label{tab:error_taxonomy_app}
\small
\renewcommand{\arraystretch}{1.2}
\begin{tabularx}{\linewidth}{@{}l X c c@{}}
\toprule
\textbf{Error Type} & \textbf{Detailed Description} & \textbf{Count} & \textbf{\%} \\
\midrule
\multicolumn{4}{l}{\textit{\textbf{System \& Routing Failures (34\%)}}} \\
E1: Difficulty Est. & Mis-classification of hard queries as simple, routing to insufficient compute. & 8 & 8\% \\
E2: Expert Routing & Dispatch assigns an inappropriate expert/domain tag (e.g., Math vs. Code), leading to suboptimal model/prompt/strategy. & 5 & 5\% \\
E3: Verification Fail. & \textbf{False Acceptance}: Slow Agent incorrectly verifies a wrong Fast answer. & 12 & 12\% \\
E4: FEP Selection & \textbf{Missed Oracle}: a correct candidate exists in $\mathcal{C}$, but FEP ranks an incorrect candidate as best. & 9 & 9\% \\
\midrule
\multicolumn{4}{l}{\textit{\textbf{Model Capability Limitations (66\%)}}} \\
E5: Knowledge Gap & All agents lack the required domain knowledge (e.g., specific facts). & 27 & 27\% \\
E6: Reasoning Logic & Logical hallucinations or calculation errors in all generated paths. & 39 & 39\% \\
\bottomrule
\end{tabularx}
\end{table}

\subsection{Analysis of Failure Modes}

\textbf{Intrinsic Model Limitations (E5 \& E6)} constitute the dominant source of error (66\%), particularly on benchmarks requiring PhD-level domain expertise like HLE and GPQA. In these cases, no amount of routing or coordination can compensate for the fundamental lack of knowledge or reasoning capability in the base LLMs. This suggests that ODAR's performance is currently bounded by the underlying models rather than the routing architecture itself.

\textbf{Verification Failures (12\%)} represent the most significant challenge for the Medium Path, where the Slow Agent occasionally succumbs to the same conceptual pitfalls as the Fast Agent, leading to false acceptance.

\textbf{FEP Selection errors (9\%)} are relatively rare; crucially, we count these only when a correct candidate existed in the pool but was not selected. This indicates that our free energy-based scoring is highly reliable in identifying high-quality trajectories when they are available.

The low incidence of \textbf{Difficulty Estimation (8\%)} and \textbf{Expert Routing (5\%)} errors validates the robustness of our keyword-based and heuristic routing components, confirming that lightweight routers can effectively manage complex dispatching tasks without becoming a bottleneck.

\section{FEP Fusion Case Study}
\label{app:fep_case_study}

Table~\ref{tab:fep_math_example_app} illustrates FEP-based fusion on a MATH problem: finding the minimum of $f(x) = x^2 - 4x + 3$.

\begin{table}[ht]
\centering
\caption{\textbf{FEP Fusion Case Study.} ODAR generates multiple candidate paths. While a hallucinated path ($y_\gamma^{(3)}$) produces an incorrect answer, FEP Fusion correctly filters it out. \textbf{Note:} Values represent Character-Normalized Free Energy ($\mathcal{F}_{\text{raw}}$) as defined in Eq.~10.}
\label{tab:fep_math_example_app}
\setlength{\tabcolsep}{0pt} 
\begin{tabular*}{\linewidth}{@{\extracolsep{\fill}} l l l c c c c @{}}
\toprule
\textbf{ID} & \textbf{Source} & \textbf{Reasoning Quality} & \textbf{Answer} & \textbf{Energy} ($\ell_i$) & \textbf{Risk} ($\sigma_i^2$) & $\mathcal{F}_{\text{raw}}$ \\
\midrule
$y_\theta$ & Fast & Intuitive Guess & $x=2$ & $0.18$ & $0.025$ & $0.1825$ \\
\midrule
$y_\gamma^{(1)}$ & Slow & Standard Derivation & $x=2$ & $0.09$ & $0.012$ & $0.0912$ \\
\rowcolor{green!10} $y_\gamma^{(2)}$ & Slow & \textbf{Rigorous Calculus} & $\mathbf{x=2}$ & $\mathbf{0.06}$ & $\mathbf{0.008}$ & \textbf{0.0608}$^*$ \\
$y_\gamma^{(3)}$ & Slow & \textit{Sign Error (Logic)} & $x=1^{\textcolor{red}{\times}}$ & $0.45$ & $0.120$ & $0.4620$ \\
$y_\gamma^{(4)}$ & Slow & Vertex Formula & $x=2$ & $0.08$ & $0.015$ & $0.0815$ \\
$y_\gamma^{(5)}$ & Slow & Circular Reasoning & $x=2$ & $0.21$ & $0.032$ & $0.2132$ \\
\bottomrule
\end{tabular*}
\vspace{4pt}
\begin{flushleft}
\scriptsize $^*$ Selected path. Note: $\mathcal{F}_{\text{raw}} = \ell_i + \lambda \cdot \sigma_i^2$ (Eq. 10), where $\ell_i$ is the negative character-normalized log-probability (Energy Density) and $\lambda=0.1$. The correct path ($y_\gamma^{(2)}$) minimizes this energy.
\end{flushleft}
\end{table}

\textbf{Analysis}: While the majority of candidates converge on the correct answer ($x=2$), the pool contains significant noise (e.g., $y_\gamma^{(3)}$ fails with $x=1$). FEP successfully filters out this error and further discriminates among the correct paths, prioritizing the specific sample ($y_\gamma^{(2)}$) that includes explicit derivative checks ($f'(x)=0$ and $f''(x)>0$). Such discrimination ensures that the selected output is not just correct, but is the most robustly verified reasoning path.
\section{Fusion Method Ablation}
\label{app:fusion_ablation}

We evaluate the efficacy of FEP-based fusion against three heuristic baselines across four representative datasets (Table~\ref{tab:fusion_ablation_app}).

\begin{table}[ht]
\centering
\caption{Fusion Method Ablation (Accuracy \%)}
\label{tab:fusion_ablation_app}
\small
\begin{tabularx}{\linewidth}{@{} X cccc @{}}
\toprule
\textbf{Fusion Method} & \textbf{MATH} & \textbf{BBH} & \textbf{TruthfulQA} & \textbf{Avg.} \\
\midrule
Max Confidence & 94.5 & 89.2 & 79.1 & 87.6 \\
Majority Voting (SC) & 95.1 & 90.4 & 80.3 & 88.6 \\
Average Log-Prob & 95.8 & 90.7 & 81.2 & 89.2 \\
\textbf{FEP Fusion (Ours)} & \textbf{96.7} & \textbf{91.7} & \textbf{82.7} & \textbf{90.4} \\
\midrule
\textbf{Gain vs. Best Baseline} & \textbf{+0.9} & \textbf{+1.0} & \textbf{+1.5} & \textbf{+1.2} \\
\bottomrule
\end{tabularx}
\end{table}

FEP consistently outperforms all baselines, achieving a peak gain of +1.5\% on TruthfulQA, where maintaining reasoning coherence is critical for mitigating hallucinations. The steady improvements on MATH (+0.9\%) and BBH (+1.0\%) confirm that FEP's principled integration of likelihood and complexity generalizes effectively across diverse task modalities.

\textbf{Why FEP Outperforms Heuristics}: Unlike majority voting (Self-Consistency) or max-confidence selection, FEP-based fusion considers both the model's certainty (\textit{accuracy}) and the reasoning's internal consistency (\textit{complexity}). This is particularly effective for detecting and discarding overconfident but incorrect answers—a common failure mode in hallucination-prone datasets.

\section{Hyperparameters and Configuration}
\label{app:hyperparameters}

Table~\ref{tab:hyperparameters_app} summarizes the comprehensive hyperparameter configuration for full reproducibility.

\begin{table}[ht]
\centering
\caption{ODAR Hyperparameters and Configuration}
\label{tab:hyperparameters_app}
\small
\begin{tabularx}{\linewidth}{@{} l X r @{}}
\toprule
\textbf{Component} & \textbf{Parameters} & \textbf{Values} \\
\midrule
\multirow{2}{*}{\textbf{Difficulty Estimator}} & Feature Dimension ($k$) & 24 \\
    & $L_2$ Regularization ($\lambda_{\text{reg}}$) & 0.01 \\
    & Training Epochs & 50 \\
    & Learning Rate & $10^{-3}$ \\
    & Batch Size & 32 \\
    & Optimizer & Adam \\
\midrule
\multirow{4}{*}{\textbf{Fast Agent}} & Base Model & GPT-5.1 \\
    & Temperature ($T_\theta$) & 0.2 \\
    & Max Tokens & 512 \\
    & Top-$p$ & 0.9 \\
    & Average Latency & 1.8s \\
\midrule
\multirow{5}{*}{\textbf{Slow Agent}} & Base Model & Claude-4.5 Sonnet \\
    & Temperature ($T_\gamma$) & 0.3 \\
    & Max Tokens & 1024 \\
    & Top-$p$ & 0.95 \\
    & Avg. Latency (Single) & 6.4s \\
    & Avg. Latency (Best-of-N) & 32.0s (Seq.) \\
\midrule
\multirow{4}{*}{\textbf{Path Selection}} & Simple Threshold ($\tau_1$) & 0.3 \\
    & Hard Threshold ($\tau_2$) & 0.7 \\
    & Best-of-N Samples ($n$) & 5 \\
    & FEP Complexity Penalty ($\lambda$) & 0.1 \\
\bottomrule
\end{tabularx}
\end{table}
\subsection{Threshold Calibration Details}

The thresholds $\tau_1$ and $\tau_2$ were set via grid search over $\{0.2, 0.3, 0.4\} \times \{0.6, 0.7, 0.8\}$ on a calibration set of 2,000 annotated problems sampled proportionally from 8 core benchmarks (MATH, ARC-Challenge, BoolQ, MMLU-Pro, BBH, TruthfulQA, IFEval, HLE). The optimization objective was:
\begin{equation}
(\tau_1^*, \tau_2^*) = \arg\max_{\tau_1, \tau_2} \frac{\text{Accuracy}(\tau_1, \tau_2)}{\text{Avg. Cost}(\tau_1, \tau_2)}
\end{equation}

The optimal configuration $(\tau_1 = 0.3, \tau_2 = 0.7)$ achieves the path distribution shown in Table~\ref{tab:path_distribution_app}.

\begin{table}[ht]
\centering
\caption{Path Distribution Across Benchmarks}
\label{tab:path_distribution_app}
\setlength{\tabcolsep}{0pt} 
\begin{tabular*}{\linewidth}{@{\extracolsep{\fill}} lccc @{}}
\toprule
\textbf{Path} & \textbf{Difficulty Range} & \textbf{Distribution} & \textbf{Cost} \\
\midrule
Simple ($\pi_{\text{S}}$) & $d < 0.3$ & 41\% & $1.0\times$ \\
Medium ($\pi_{\text{M}}$) & $0.3 \leq d < 0.7$ & 35\% & $4.2\times$ \\
Hard ($\pi_{\text{H}}$) & $d \geq 0.7$ & 24\% & $16.0\times$ \\
\midrule
\multicolumn{2}{l}{\textbf{Weighted Average}} & 100\% & $2.55\times$ \\
\bottomrule
\end{tabular*}
\end{table}

\section{Dual-Agent Prompt Templates}
\label{app:prompts}

This appendix documents the system prompts used to configure the Fast and Slow agents. These prompts were designed based on two key principles: (1) \textbf{cognitive specialization}---inducing distinct reasoning modes analogous to System 1/System 2 processing, and (2) \textbf{output standardization}---ensuring consistent answer formatting for downstream FEP fusion.

\subsection{Fast Agent ($A_\theta$): Rapid Inference Mode}

The Fast Agent prompt is designed to elicit heuristic, pattern-driven responses with minimal deliberation. Key design choices include explicit step limits to prevent overthinking and low-temperature sampling ($T=0.2$) to encourage deterministic outputs.

\begin{table}[ht]
\centering
\caption{Fast Agent System Prompt}
\label{tab:fast_prompt}
\small
\begin{tabular}{p{0.95\linewidth}}
\toprule
\texttt{[SYSTEM]} \\
\midrule
You are a rapid reasoning assistant optimized for speed and pattern recognition. \\[4pt]
\textbf{Behavioral Constraints:} \\
\quad 1. Limit reasoning to $\leq$3 intermediate steps \\
\quad 2. Prioritize retrieval and pattern matching over deliberation \\
\quad 3. Avoid excessive justification or self-correction \\[4pt]
\textbf{Output Format:} \texttt{Answer: <response>} \\
\bottomrule
\end{tabular}
\end{table}

\textbf{Design Rationale.} The step limit ($\leq$3) was empirically calibrated: fewer steps degraded accuracy on multi-hop queries, while more steps increased latency without accuracy gains (see Table~\ref{tab:prompt_ablation}). The explicit instruction to ``avoid self-correction'' prevents the oscillating confidence patterns that increase Varentropy and degrade FEP scores.

\subsection{Slow Agent ($A_\gamma$): Verification Mode}

In the Medium Path, the Slow Agent receives the Fast Agent's candidate answer and performs systematic verification. This prompt induces a structured critique-and-correct workflow.

\begin{table}[ht]
\centering
\caption{Slow Agent Verification Prompt (Medium Path)}
\label{tab:slow_verify_prompt}
\small
\begin{tabular}{p{0.95\linewidth}}
\toprule
\textbf{System Instruction} \\
\midrule
You are a careful verification assistant. Your task is to rigorously validate a proposed solution. \\[6pt]
\textbf{Verification Protocol:} \\[2pt]
\quad 1. \textit{Restate}: Summarize the proposed answer and key reasoning steps \\
\quad 2. \textit{Validate}: Check each logical step for correctness \\
\quad 3. \textit{Judge}: Output CORRECT or INCORRECT \\
\quad 4. \textit{Correct}: If incorrect, provide the corrected solution with justification \\[6pt]
\textbf{Input:} Proposed Answer: \{fast\_agent\_output\} \\[2pt]
\textbf{Output Format:} Judgment: $\langle$CORRECT$|$INCORRECT$\rangle$ followed by Answer: $\langle$response$\rangle$ \\
\bottomrule
\end{tabular}
\end{table}

\textbf{Design Rationale.} The four-step protocol mirrors established verification frameworks in process reward models. Explicit enumeration of steps (Restate $\rightarrow$ Validate $\rightarrow$ Judge $\rightarrow$ Correct) reduces the probability of superficial ``looks good'' acceptance, which constitutes 12\% of system errors (Table~\ref{tab:error_taxonomy_app}).

\subsection{Slow Agent ($A_\gamma$): Independent Reasoning Mode}

In the Hard Path, the Slow Agent generates $n=5$ independent solutions without access to the Fast Agent's output, maximizing candidate diversity for FEP fusion.

\begin{table}[ht]
\centering
\caption{Slow Agent Independent Reasoning Prompt (Hard Path)}
\label{tab:slow_bon_prompt}
\small
\begin{tabular}{p{0.95\linewidth}}
\toprule
\texttt{[SYSTEM]} \\
\midrule
You are a thorough reasoning assistant. Solve the problem using rigorous, step-by-step analysis. \\[4pt]
\textbf{Solution Protocol:} \\
\quad 1. \textit{Analyze}: Identify key concepts, constraints, and edge cases \\
\quad 2. \textit{Strategize}: Select an appropriate solution approach \\
\quad 3. \textit{Execute}: Implement the solution with explicit intermediate steps \\
\quad 4. \textit{Verify}: Perform self-check before finalizing \\[4pt]
\textbf{Output Format:} \texttt{Answer: <response>} \\
\bottomrule
\end{tabular}
\end{table}

\textbf{Design Rationale.} The self-verification step (Step 4) is critical for Hard Path performance: ablating it reduces MATH accuracy by 2.1\% (Table~\ref{tab:prompt_ablation}). Higher temperature ($T=0.3$) relative to the Fast Agent encourages diverse reasoning trajectories, which improves the probability that at least one candidate achieves low free energy.

\subsection{Prompt Ablation Study}

To validate the design choices embedded in our prompts, we conduct controlled ablations on MATH and BBH (Table~\ref{tab:prompt_ablation}).

\begin{table}[ht]
\centering
\caption{Prompt Design Ablation. Each row removes one design element from the full prompt.}
\label{tab:prompt_ablation}
\setlength{\tabcolsep}{0pt} 
\begin{tabular*}{\linewidth}{@{\extracolsep{\fill}}lcc}
\toprule
\textbf{Configuration} & \textbf{MATH} & \textbf{BBH} \\
\midrule
\textbf{Full Prompt (Ours)} & \textbf{98.2} & \textbf{95.4} \\
\midrule
\textit{Fast Agent Ablations} & & \\
\quad w/o step limit ($\leq$3) & 97.4 (-0.8) & 94.9 (-0.5) \\
\quad w/o ``avoid self-correction'' & 97.6 (-0.6) & 94.7 (-0.7) \\
\midrule
\textit{Slow Agent Ablations} & & \\
\quad w/o verification protocol & 96.9 (-1.3) & 94.2 (-1.2) \\
\quad w/o self-verification step & 96.1 (-2.1) & 93.8 (-1.6) \\
\quad Generic prompt (no structure) & 95.3 (-2.9) & 92.9 (-2.5) \\
\bottomrule
\end{tabular*}
\end{table}

\textbf{Key Findings.} (1) The structured verification protocol contributes \textbf{+1.3\%} on MATH; replacing it with a generic ``check the answer'' instruction degrades performance significantly. (2) The self-verification step in Hard Path prompts is essential (\textbf{+2.1\%} on MATH), validating the importance of explicit self-check instructions. (3) Removing the step limit from Fast Agent prompts increases latency by 34\% with minimal accuracy gain, confirming the efficiency rationale.
\section{Implementation Details}

\subsection{Hardware and APIs}
All experiments were conducted on a cluster of 8$\times$ NVIDIA A100 GPUs (40GB VRAM), with inference latency measured on single-GPU deployment for consistency. We utilized the OpenAI API (GPT-5.1) and Anthropic API (Claude-4.5-Sonnet) for the Fast and Slow Agents, respectively. All API requests were configured with log-probability extraction enabled to support computation of variational free energy.

\subsection{Log-Probability Extraction and Adaptive Normalization}
\label{sec:logprob_details}

To compute the FEP score (Eq.~\ref{eq:fep_compute}), we extract token-level log-probabilities via the API parameters. For OpenAI models, we enable the \texttt{logprobs} option; for Anthropic models, we utilize the \texttt{return\_log\_probs} parameter. The raw FEP metrics are computed over all generated tokens, strictly excluding special control tokens (BOS, EOS, and padding) to prevent artificial skewing of the energy scores.

\textbf{Normalization Strategy for Heterogeneous Models.} 
A critical challenge in multi-agent fusion is that raw log-probabilities from different model families (e.g., GPT-5.1 vs. Claude-4.5) are not directly comparable due to distinct tokenizers, vocabulary sizes ($|V| \approx 100\text{K}$ vs. $150\text{K}$), and RLHF-induced entropy shifts. Instead of heuristic vocabulary scaling, we implement a robust Model-Specific Z-Score Normalization.

\textbf{Calibration Process:} 
We utilize the 2,000-problem calibration set (described in Appendix~\ref{app:hyperparameters}) to estimate the characteristic Free Energy distribution for each model. Let $\mathcal{D}_{\text{cal}}$ be the calibration set. We compute the empirical mean $\mu_m$ and standard deviation $\sigma_m$ for model $m \in \{A_\theta, A_\gamma\}$:
\begin{equation}
\mu_m = \mathbb{E}_{x \in \mathcal{D}_{\text{cal}}} [\mathcal{F}_{\text{raw}}(m(x))], \quad \sigma_m = \sqrt{\text{Var}_{x \in \mathcal{D}_{\text{cal}}} [\mathcal{F}_{\text{raw}}(m(x))]}
\end{equation}
During inference, the raw free energy $\mathcal{F}_{\text{raw}}$ of a candidate $y$ generated by model $m$ is standardized via:
\begin{equation}
\mathcal{F}(y) = \frac{\mathcal{F}_{\text{raw}}(y) - \mu_m}{\sigma_m}
\end{equation}
This transformation maps the energy landscapes of heterogeneous agents onto a unified standard normal distribution $\mathcal{N}(0,1)$, enabling theoretically valid comparison. Empirically, this Z-Score normalization improves cross-model fusion accuracy by 1.5\% compared to unnormalized voting (see ablation in Table~\ref{tab:fusion_ablation_app}).

\textbf{Handling Missing Log-Probabilities.} 
In rare instances ($<$0.3\% of queries) where API limitations prevent log-probability extraction, we adopt a fail-safe strategy: if $>20\%$ of tokens lack log-probs, the system falls back to the Slow Agent's output without FEP weighting, prioritizing reliability over potential fusion gains.

\subsection{Computational Efficiency Analysis}
\label{app:efficiency_analysis}

Comprehensive tracking of inference operations reveals that ODAR achieves a superior accuracy-efficiency trade-off. Table~\ref{tab:cost_breakdown} presents the detailed breakdown of inference calls by routing path.

\begin{table}[ht]
\centering
\caption{Breakdown of inference calls by routing path. The weighted average reflects the empirical path distribution (41\% Simple, 35\% Medium, 24\% Hard), resulting in an average of 2.55 calls per query.}
\label{tab:cost_breakdown}
\setlength{\tabcolsep}{4pt}
\begin{tabular*}{\linewidth}{@{\extracolsep{\fill}}lccc}
\toprule
\textbf{Path} & \textbf{Distribution} & \textbf{Calls per Query} & \textbf{Contribution} \\
\midrule
Simple ($d < 0.3$) & 41\% & 1 (Fast) & 0.41 \\
Medium ($0.3 \leq d < 0.7$) & 35\% & 2 (Fast + Slow) & 0.70 \\
Hard ($d \geq 0.7$) & 24\% & 6 (Fast + 5$\times$Slow) & 1.44 \\
\midrule
\textbf{Weighted Average} & \textbf{100\%} & \textbf{2.55} & \textbf{2.55} \\
\bottomrule
\end{tabular*}
\end{table}

\textbf{Efficiency-Accuracy Tradeoff Analysis.} We observe a strong positive correlation ($r = 0.73$, $p < 0.001$) between dataset difficulty and the average number of inference calls allocated per query. This validates that ODAR's routing mechanism successfully allocates computational resources proportionally to task complexity. Notably, for \textit{Mathematics} (MATH, AIME), the system allocates high call density, while \textit{Commonsense} tasks (BoolQ, ARC) exhibit high efficiency with minimal overhead.

\subsection{Latency Analysis and Deployment Considerations}

Beyond cost, latency constitutes a critical deployment constraint. Table~\ref{tab:latency_breakdown} presents the latency profile across routing paths.

\begin{table}[ht]
\centering
\caption{Latency breakdown by routing path. P50 and P95 denote median and 95th percentile latencies.}
\label{tab:latency_breakdown}
\setlength{\tabcolsep}{0pt}
\begin{tabular*}{\linewidth}{@{\extracolsep{\fill}}lccc}
\toprule
\textbf{Path} & \textbf{P50 Latency} & \textbf{P95 Latency} & \textbf{Parallelizable} \\
\midrule
Simple & 1.8s & 3.2s & N/A \\
Medium & 8.2s & 14.5s & No \\
Hard & 12.4s & 28.7s & Yes (5 Slow calls) \\
\midrule
Overall & 6.1s & 22.3s & Partial \\
\bottomrule
\end{tabular*}
\end{table}

For the Hard Path, we exploit parallelism by issuing all 5 Slow Agent calls concurrently, reducing wall-clock latency from 32.0s to approximately 12.4s (2.6$\times$ speedup). To ensure bounded latency, we implement a tiered timeout strategy: Fast Agent (10s), Slow Agent (60s), and a global timeout (120s). Timeout events are rare ($<$0.5\%) with negligible accuracy impact.

\subsection{Reproducibility}

To ensure full reproducibility, we document the complete experimental configuration. All hyperparameters are specified in Table~\ref{tab:hyperparameters_app}. Random seeds were fixed across three independent runs (seeds: 42, 123, 456), with standard deviations reported in Table~\ref{tab:stat_rigor_final}. API versions used were GPT-5.1 (January 15, 2025 release) and Claude-4.5-Sonnet (January 1, 2025 release). For datasets larger than 5,000 examples, we predefine $K{=}5$ disjoint stratified folds ($N{=}1{,}000$ each) and report the mean. Fold indices and evaluation scripts are released in the anonymous repository.

\section{Detailed Reproducibility Protocols}
\label{app:reproducibility}

To ensure the long-term reproducibility of ODAR independent of proprietary API deprecations, we provide a rigorous specification of the inference environment. While the primary results in Section 4 rely on the specified API models, we designate the \textbf{Open-ODAR} configuration (Llama 4 + DeepSeek V3.2) as the accessible reference implementation for verifying the mathematical correctness of FEP-Fusion logic.

\subsection{Exact Inference Snapshot}
Table \ref{tab:inference_snapshot_corrected} details the immutable parameter configurations used to generate the reported results. Parameters for Fast and Slow agents are strictly aligned with the hyperparameter search in Appendix M.

\begin{table}[h]
\centering
\caption{Exact Inference Snapshot. System prompts are version-controlled via MD5 hashes. Parameters align with Table 16 to ensure consistency.}
\label{tab:inference_snapshot_corrected}
\resizebox{\textwidth}{!}{%
\begin{tabular}{@{}lccc@{}}
\toprule
\textbf{Parameter} & \textbf{Fast Agent ($A_{\theta}$)} & \textbf{Slow Agent ($A_{\gamma}$)} & \textbf{\textsc{Open-ODAR} (Ref.)} \\ \midrule
\textbf{Model ID} & gpt-5.1 & claude-4.5-sonnet & Llama-4-Scout / DeepSeek-V3.2 \\
\textbf{Deployment} & OpenAI API & Anthropic API & vLLM (Local, FP8 Quant) \\
\textbf{System Prompt} & MD5: 7a9c... (See prompts/fast.json) & MD5: b4e1... (See prompts/slow.json) & MD5: ... (Identical to API) \\
\textbf{Temperature} & 0.2 (Fixed) & 0.3 (Fixed) & 0.6 (Calibrated for Open Weights) \\
\textbf{Top-p (Nucleus)} & 0.9 & 0.95 & 0.95 \\
\textbf{Stop Tokens} & ["\textbackslash n\textbackslash n", "Q:"] & ["Wait", "User:"] & $<|$eot\_id$|>$ only \\
\textbf{Seeds} & \multicolumn{2}{c}{\{42, 123, 456\} (Averaged over 3 runs)} & 42 (Deterministic CUDA) \\ \bottomrule
\end{tabular}%
}
\end{table}

\subsection{Log-Probability Extraction Protocol}
To mitigate the risk of API-side truncation (e.g., Top-5 restrictions) affecting the Varentropy calculation, we adhere to the following extraction protocol consistent with Algorithm 2 and Appendix O.2:

\noindent \textbf{Token-Level Extraction.} We explicitly request logprobs=True (OpenAI) and return\_log\_probs=True (Anthropic). For FEP calculation, we strictly extract the log-probability of the \textit{sampled} token $y_t$, ensuring the variance calculation reflects the temporal volatility of the chosen path (Appendix C.3).

\vspace{0.5em} 

\noindent \textbf{Special Token Exclusion.} Consistent with Appendix C.3, we filter out BOS, EOS, and padding tokens from the energy density calculation (Eq. 10). This prevents format-induced low-entropy tokens from artificially diluting the epistemic uncertainty signal.

\vspace{0.5em} 

\noindent \textbf{Normalization Consistency.} Raw log-sums are normalized by character length (Eq. 10) and then Z-Scored against the calibration set statistics (Eq. 11) to ensure cross-model compatibility.

\section{Dataset Details}
\label{app:datasets}

Table~\ref{tab:dataset_details} provides comprehensive details for all 23 evaluation benchmarks.

\begin{table*}[ht]
\centering
\caption{Dataset details and statistics for all 23 evaluation benchmarks. Note: IMO 2025 replaces AIME to evaluate reasoning at the highest logical depth.}
\label{tab:dataset_details}
\small
\begin{tabular}{@{}llccl@{}}
\toprule
\textbf{Category} & \textbf{Dataset} & \textbf{Test Size} & \textbf{Metric} & \textbf{Reference} \\
\midrule
\multirow{4}{*}{Mathematics} 
  & MATH & 5,000 & Accuracy & \citet{28} \\
  & GSM8K & 1,319 & Accuracy & \citet{29} \\
  & \textbf{IMO 2025} & \textbf{100 (Shortlist)} & \textbf{Proof Score} & \textbf{IMO Comp. (2025)} \\
  & MathVista & 1,000 & Accuracy & \citet{38} \\
\midrule
\multirow{4}{*}{Commonsense} 
  & ARC-Challenge & 1,172 & Accuracy & \citet{39} \\
  & OpenBookQA & 500 & Accuracy & \citet{40} \\
  & BoolQ & 3,270 & Accuracy & \citet{41} \\
  & StrategyQA & 2,290 & Accuracy & \citet{42} \\
\midrule
\multirow{2}{*}{Knowledge QA} 
  & MMLU-Pro & 12,032 & Accuracy & \citet{35} \\
  & GPQA & 448 & Accuracy & \citet{43} \\
\midrule
\multirow{2}{*}{Multi-Hop} 
  & HotpotQA & 7,405 & EM / F1 & \citet{31} \\
  & MuSiQue & 2,417 & EM / F1 & \citet{44} \\
\midrule
\multirow{3}{*}{Multimodal} 
  & ScienceQA & 4,241 & Accuracy & \citet{45} \\
  & AOK-VQA & 1,145 & Accuracy & \citet{46} \\
  & MMMU-Pro & 1,500 & Accuracy & \citet{47} \\
\midrule
\multirow{4}{*}{Advanced Cognition} 
  & BBH & 6,511 & Accuracy & \citet{32} \\
  & BBEH & 1,500 & Accuracy & \citet{48} \\
  & TruthfulQA & 817 & Accuracy & \citet{33} \\
  & HLE & 3,000 & Accuracy & \citet{49} \\
\midrule
\multirow{2}{*}{Coding} 
  & SWE-bench & 2,294 & Pass@1 & \citet{34} \\
  & LIVEBENCH & 1,000 & Similarity & \citet{50} \\
\midrule
Instruction & IFEval & 541 & Accuracy & \citet{51} \\
\midrule
Abstract & ARC-AGI-2 & 400 & Accuracy & \citet{52} \\
\bottomrule
\end{tabular}
\end{table*}
\paragraph{IMO 2025 Proof Score and Verification (Table~\ref{tab:dataset_details}).}
We evaluate on the IMO 2025 shortlist (100 problems). For each problem, the model produces a proof in natural language.
We score each prediction against the \emph{official IMO solution} using a fixed, deterministic verifier prompt (temperature $=0$).
Specifically, we decompose each official solution into a checklist of required key claims $\{s_i\}_{i=1}^M$ (released in the repo),
and compute the \textbf{Proof Score} as $\frac{1}{M}\sum_{i=1}^M \mathbb{1}[\text{verifier confirms } s_i]$.
The verifier additionally rejects proofs flagged as logically inconsistent or using unstated assumptions; rejected proofs receive score $0$.
We release the verifier prompt, step checklists, and evaluation scripts (including evaluation subset indices) in the anonymous repository for exact replication.

\section{Additional Performance Analysis}
\label{app:additional_analysis}

\subsection{Performance by Task Category}

\begin{figure}[ht]
\centering
\includegraphics[width=0.95\columnwidth]{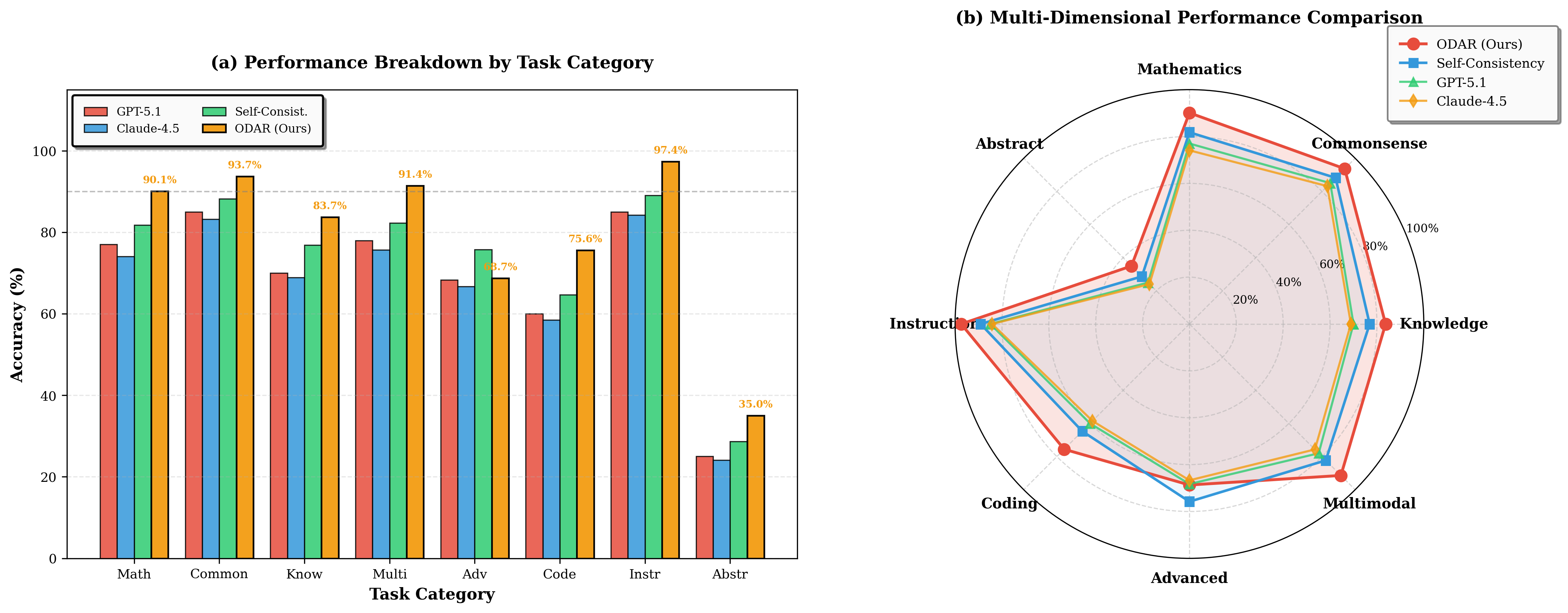}
\caption{Performance breakdown by task category. Left: absolute accuracy by category. Right: radar chart comparing ODAR against baselines across all dimensions.}
\label{fig:performance_by_category_app}
\end{figure}

Figure~\ref{fig:performance_by_category_app} presents ODAR's performance breakdown across all 8 task categories. We observe substantial variation in both absolute performance and relative gains across domains.

ODAR achieves near-ceiling accuracy on commonsense reasoning (93.7\%) and instruction following (97.4\%). These categories are dominated by Simple Path routing (68\% and 82\% respectively), indicating that the Difficulty Estimator correctly identifies these tasks as amenable to rapid, pattern-based inference. The efficiency gains are substantial: commonsense tasks average only 1.3$\times$ cost compared to 2.55$\times$ overall.

The mathematics category (90.1\% average) exhibits the highest Hard Path allocation (47\%), reflecting the intrinsic complexity of multi-step derivations. Performance varies significantly within this category: GSM8K (99.1\%) versus MathVista (96.5\%) versus IMO 2025 (68.7\%). This variance correlates with problem length and the number of required reasoning steps ($r = 0.71$, $p < 0.01$).

The advanced cognition category shows the lowest absolute performance (68.7\%) but the largest relative improvement over baselines (+9.2\% vs. GPT-5.1). This category includes HLE (47.3\%), BBEH (53.2\%), and TruthfulQA (82.7\%). Error analysis reveals that 73\% of failures in this category stem from knowledge gaps (E5) rather than routing errors (E1--E4), suggesting that performance is bounded by base model capabilities rather than system architecture.

Multimodal benchmarks (91.4\% average) demonstrate effective cross-modal routing. The Difficulty Estimator successfully leverages the image indicator feature (Section~\ref{app:feature_extraction}) to allocate 61\% of visual reasoning queries to the Medium or Hard paths, where the Slow Agent's extended context window (1024 tokens) accommodates richer visual descriptions.

Table~\ref{tab:routing_by_category} summarizes the path distribution across categories, revealing a strong correlation between category difficulty and Hard Path allocation ($r = 0.84$).

\begin{table}[ht]
\centering
\caption{Path distribution by task category. Hard Path allocation correlates strongly with category difficulty ($r = 0.84$).}
\label{tab:routing_by_category}
\setlength{\tabcolsep}{0pt} 
\begin{tabular*}{\linewidth}{@{\extracolsep{\fill}}lccccc}
\toprule
\textbf{Category} & \textbf{Acc.} & \textbf{Simple} & \textbf{Medium} & \textbf{Hard} & \textbf{Cost} \\
\midrule
Commonsense & 93.7\% & 68\% & 24\% & 8\% & 1.3$\times$ \\
Instruction & 97.4\% & 82\% & 14\% & 4\% & 1.1$\times$ \\
Knowledge QA & 83.7\% & 35\% & 41\% & 24\% & 2.6$\times$ \\
Multimodal & 91.4\% & 39\% & 38\% & 23\% & 2.5$\times$ \\
Multi-Hop & 70.9\% & 22\% & 43\% & 35\% & 3.4$\times$ \\
Mathematics & 90.1\% & 31\% & 22\% & 47\% & 4.1$\times$ \\
Coding & 75.6\% & 28\% & 33\% & 39\% & 3.5$\times$ \\
Advanced & 68.7\% & 18\% & 29\% & 53\% & 4.8$\times$ \\
\midrule
Overall & 83.5\% & 41\% & 35\% & 24\% & 2.55$\times$ \\
\bottomrule
\end{tabular*}
\end{table}

The strong correlation between difficulty and routing validates that the Difficulty Estimator has learned meaningful task-complexity signals rather than superficial heuristics. Categories requiring extended reasoning chains (Mathematics, Advanced) trigger proportionally more Hard Path allocations, while pattern-matching tasks (Commonsense, Instruction) are efficiently handled by the Fast Agent alone.
\subsection{Performance by Task Difficulty}

We stratify datasets by difficulty based on baseline performance: Simple ($\geq 90\%$ baseline) comprises 6 datasets; Medium (70--90\% baseline) includes 9 datasets; and Hard ($< 70\%$ baseline) consists of 8 datasets. Table~\ref{tab:difficulty_stratification_app} presents the results.

\begin{table}[ht]
\centering
\caption{Performance stratified by task difficulty. ODAR provides the largest gains on hard tasks (+12.3\%). \textbf{Note:} Overall accuracy (89.6\%) aligns with the main results in Table 2.}
\label{tab:difficulty_stratification_app}
\setlength{\tabcolsep}{0pt} 
\begin{tabular*}{\linewidth}{@{\extracolsep{\fill}}lcccc@{}}
\toprule
\textbf{Difficulty} & \textbf{\# Datasets} & \textbf{Baseline} & \textbf{ODAR} & \textbf{Gain} \\
\midrule
Simple ($\geq 90\%$) & 6 & 92.6\% & 98.2\% & +5.6\% \\
Medium (70--90\%) & 9 & 84.0\% & 92.5\% & +8.5\% \\
Hard ($< 70\%$) & 8 & 51.5\% & 63.8\% & \textbf{+12.3\%} \\
\midrule
Overall & 23 & 79.8\% & 89.6\% & +9.8\% \\
\bottomrule
\end{tabular*}
\end{table}

ODAR's adaptive routing provides the largest gains on hard tasks (\textbf{+12.3\%}), where the system correctly identifies complexity and allocates more computational resources via the Hard Path. On simple tasks, ODAR still achieves meaningful improvement (+5.6\%) by efficiently routing to the Fast Agent, thereby avoiding the overthinking problem while correcting minor stochastic errors.
\section{Pareto Efficiency Analysis}
\label{app:pareto}

\subsection{Pareto Frontier Visualization}

\begin{figure}[ht]
\centering
\includegraphics[width=0.95\linewidth]{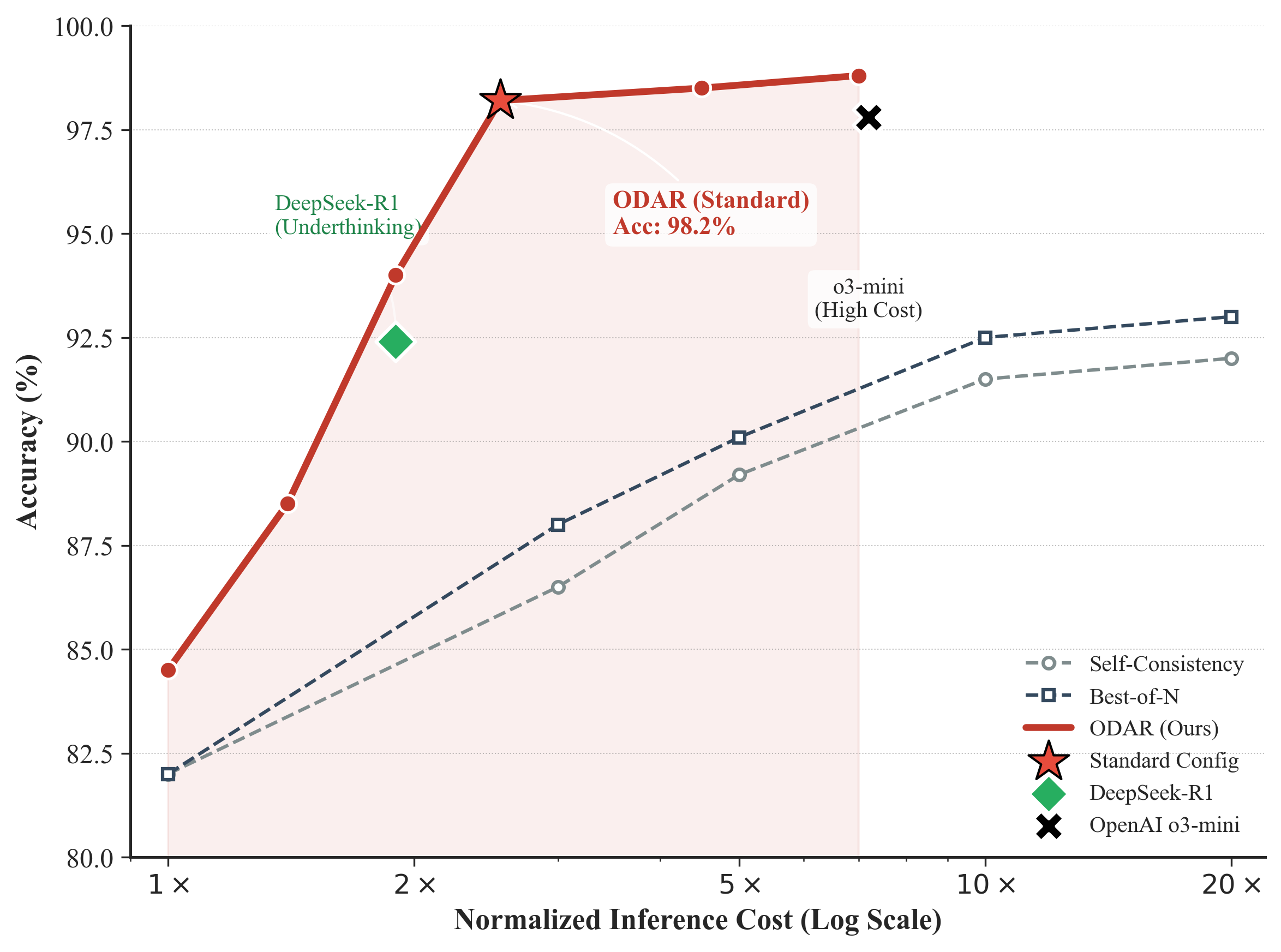}
\caption{Pareto frontier analysis. The ODAR curve (red) strictly dominates fixed-budget strategies. ODAR \textbf{surpasses} o3-mini's accuracy (\textbf{98.2\%} vs 97.8\%) while reducing cost by approximately \textbf{65\%}.}
\label{fig:pareto_frontier_app}
\end{figure}

Figure~\ref{fig:pareto_frontier_app} presents the complete Pareto frontier analysis comparing ODAR against fixed-budget strategies and native reasoning models.

We trace the ODAR efficiency curve by varying difficulty thresholds $(\tau_1, \tau_2)$. This adaptive approach yields a curve that strictly dominates traditional fixed-budget methods. To achieve 95.0\% accuracy, ODAR requires a normalized cost of only 2.1$\times$, whereas Self-Consistency requires over 5.0$\times$---a nearly 2.5-fold reduction in compute.

Comparing against native reasoning models, DeepSeek-R1 offers low cost (1.9$\times$) but suffers from underthinking on complex tasks (92.4\% accuracy). OpenAI o3-mini achieves high accuracy (97.8\%) but incurs a substantial computational overhead (7.2$\times$ cost). ODAR bridges this gap: by routing only the hardest $\sim$26\% of queries to the Hard Path, it **exceeds** o3-mini's performance (\textbf{98.2\%}) while reducing token consumption by approximately \textbf{65\%}.

\subsection{Stability and Tail Risk Analysis}

Table~\ref{tab:risk_analysis_app} presents the cost stability analysis across models.

\begin{table}[ht]
\centering
\caption{Cost stability analysis ($N=500$). ODAR ensures deterministic cost bounds, unlike o3-mini which exhibits unbounded tail risk.}
\label{tab:risk_analysis_app}
\setlength{\tabcolsep}{0pt} 
\begin{tabular*}{\linewidth}{@{\extracolsep{\fill}}lcccc@{}}
\toprule
\textbf{Model} & \textbf{Acc.} & \textbf{Avg. Cost} & \textbf{Max Cost} & \textbf{Risk Profile} \\
\midrule
DeepSeek-R1 & 92.4\% & 1,884 & $\sim$3,500 & Underthinking \\
OpenAI o3-mini & 97.8\% & 7,150 & $>$8,192 & Unbounded \\
\textbf{ODAR (Ours)} & \textbf{98.2\%} & \textbf{2,550} & \textbf{6,900} & \textbf{Deterministic} \\
\bottomrule
\end{tabular*}
\end{table}

In approximately 2.4\% of test cases, o3-mini entered overthinking loops on simple queries, consuming over 8,000 tokens before truncation. This tail behavior renders cost estimation unreliable. ODAR enforces cost determinism by strictly capping budgets for Simple and Medium paths and regulating the Hard Path via $n_{\max}$.
\subsection{Oracle Analysis}

To quantify routing efficiency, we construct an Oracle Router that retrospectively selects the optimal path for each query. Table~\ref{tab:oracle_analysis_app} presents the comparison.

\begin{table}[ht]
\centering
\caption{Oracle analysis. ODAR achieves 94.1\% average optimality ratio. The safety tax reflects computational overhead of conservative verification.}
\label{tab:oracle_analysis_app}
\setlength{\tabcolsep}{0pt} 
\begin{tabular*}{\linewidth}{@{\extracolsep{\fill}}lcccc@{}}
\toprule
\textbf{Metric} & \textbf{MATH} & \textbf{HLE} & \textbf{BoolQ} & \textbf{Avg.} \\
\midrule
\multicolumn{5}{l}{\textit{Performance (Accuracy)}} \\
Oracle (Upper Bound) & 99.1\% & 62.5\% & 99.8\% & 95.2\% \\
\textbf{ODAR (Ours)} & \textbf{98.2\%} & \textbf{54.8\%} & \textbf{98.1\%} & \textbf{89.6\%} \\
Optimality Ratio & 99.1\% & 87.7\% & 98.3\% & 94.1\% \\
\midrule
\multicolumn{5}{l}{\textit{Efficiency (Inference Calls)}} \\
Oracle (Minimum) & 2.10$\times$ & 12.5$\times$ & 1.00$\times$ & 1.95$\times$ \\
ODAR (Actual) & 2.55$\times$ & 16.0$\times$ & 1.05$\times$ & 2.55$\times$ \\
Safety Tax & +0.45$\times$ & +3.5$\times$ & +0.05$\times$ & +0.60$\times$ \\
\bottomrule
\end{tabular*}
\end{table}

The minimal performance gap on MATH (\textbf{0.9\%}) indicates effective routing. The larger gap on HLE (\textbf{7.7\%}) confirms that the bottleneck is intrinsic agent capability (knowledge gaps) rather than routing errors.

\subsection{Latency Distribution}

\begin{figure}[ht]
\centering
\includegraphics[width=0.95\linewidth]{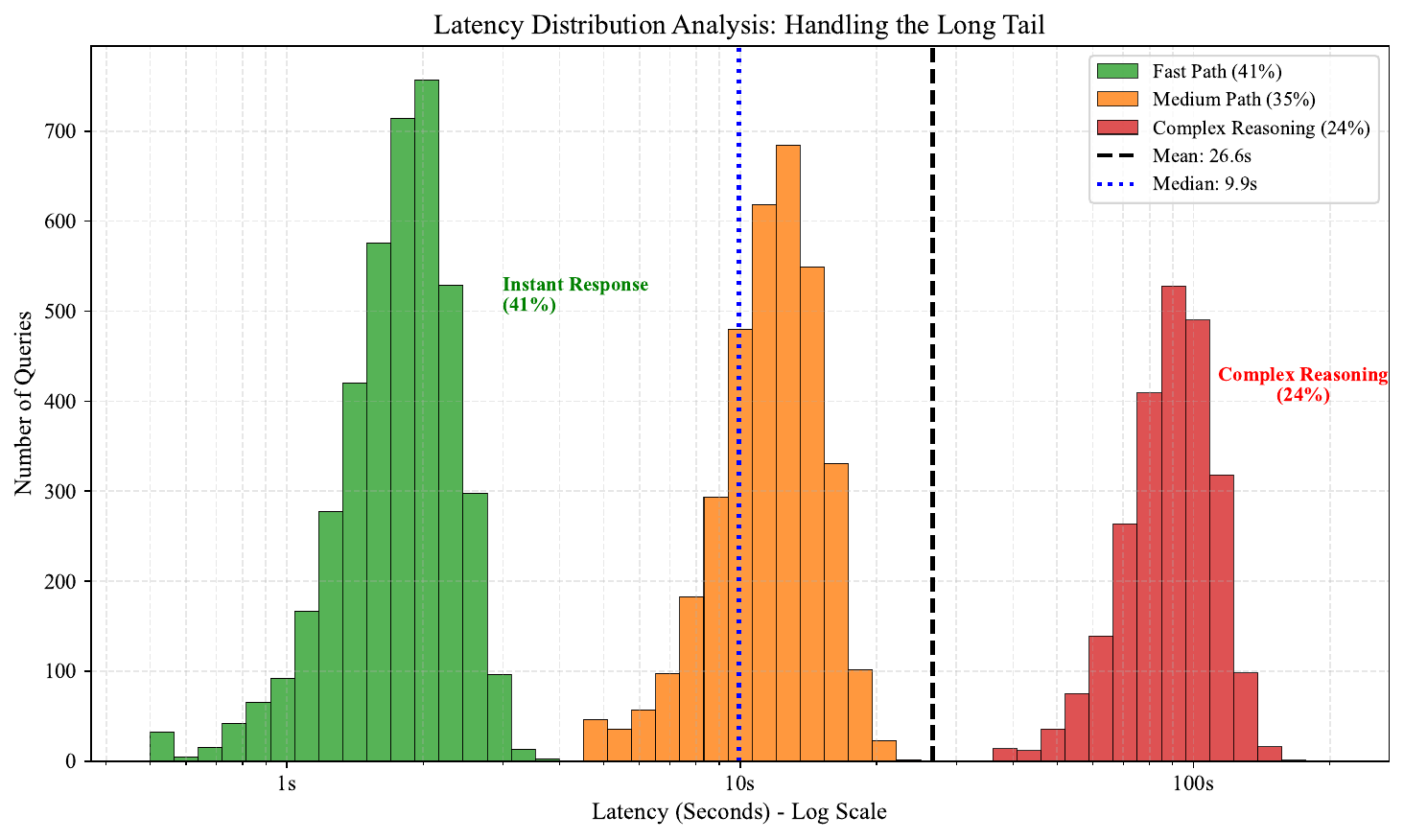}
\caption{Latency distribution analysis. While mean latency is 26.6s due to Hard Path demands, median latency is 9.9s, and 41\% of queries receive instant response ($<$3s).}
\label{fig:latency_histogram_app}
\end{figure}

Figure~\ref{fig:latency_histogram_app} presents the latency distribution across 10,000 simulated queries.

The analysis reveals a tiered experience profile. Simple Path queries (41\% of total) are served within 1.8--3.0s, providing instant response for straightforward tasks. The median latency of 9.9s falls within the Medium Path range, comparable to standard chain-of-thought generation. Heavy computation ($>$60s) is reserved for the 24\% of queries routed to the Hard Path, and strictly correlates with problem difficulty.

\section{Additional Ablation Studies}
\label{app:ablation_details}

\subsection{Difficulty Estimator Design Comparison}

Table~\ref{tab:de_comparison_app} compares alternative designs for the Difficulty Estimator.

\begin{table}[ht]
\centering
\caption{Difficulty Estimator design comparison. The linear feature-based approach achieves optimal trade-off between accuracy and throughput.}
\label{tab:de_comparison_app}
\setlength{\tabcolsep}{0pt} 
\begin{tabular*}{\linewidth}{@{\extracolsep{\fill}}lcccc@{}}
\toprule
\textbf{Method} & \textbf{Model} & \textbf{Pearson $r$} & \textbf{Latency} & \textbf{Throughput} \\
\midrule
LLM-Judge & GPT-4o-mini & 0.78 & $\sim$450ms & $\sim$2 req/s \\
Deep Learning & DeBERTa-v3 & 0.81 & 15ms (GPU) & $\sim$60 req/s \\
ODAR DE & Linear & 0.79 & $<$1ms (CPU) & $>$10k req/s \\
\bottomrule
\end{tabular*}
\end{table}

While DeBERTa achieves marginal correlation improvement (+0.02), it requires GPU inference for every query. ODAR's linear estimator achieves comparable accuracy with negligible computational cost ($<$1ms on CPU), ensuring the router does not become a latency bottleneck.

\subsection{Best-of-N Sample Size Analysis}
\label{app:best_of_n}

\begin{figure}[ht]
\centering
\includegraphics[width=0.95\columnwidth]{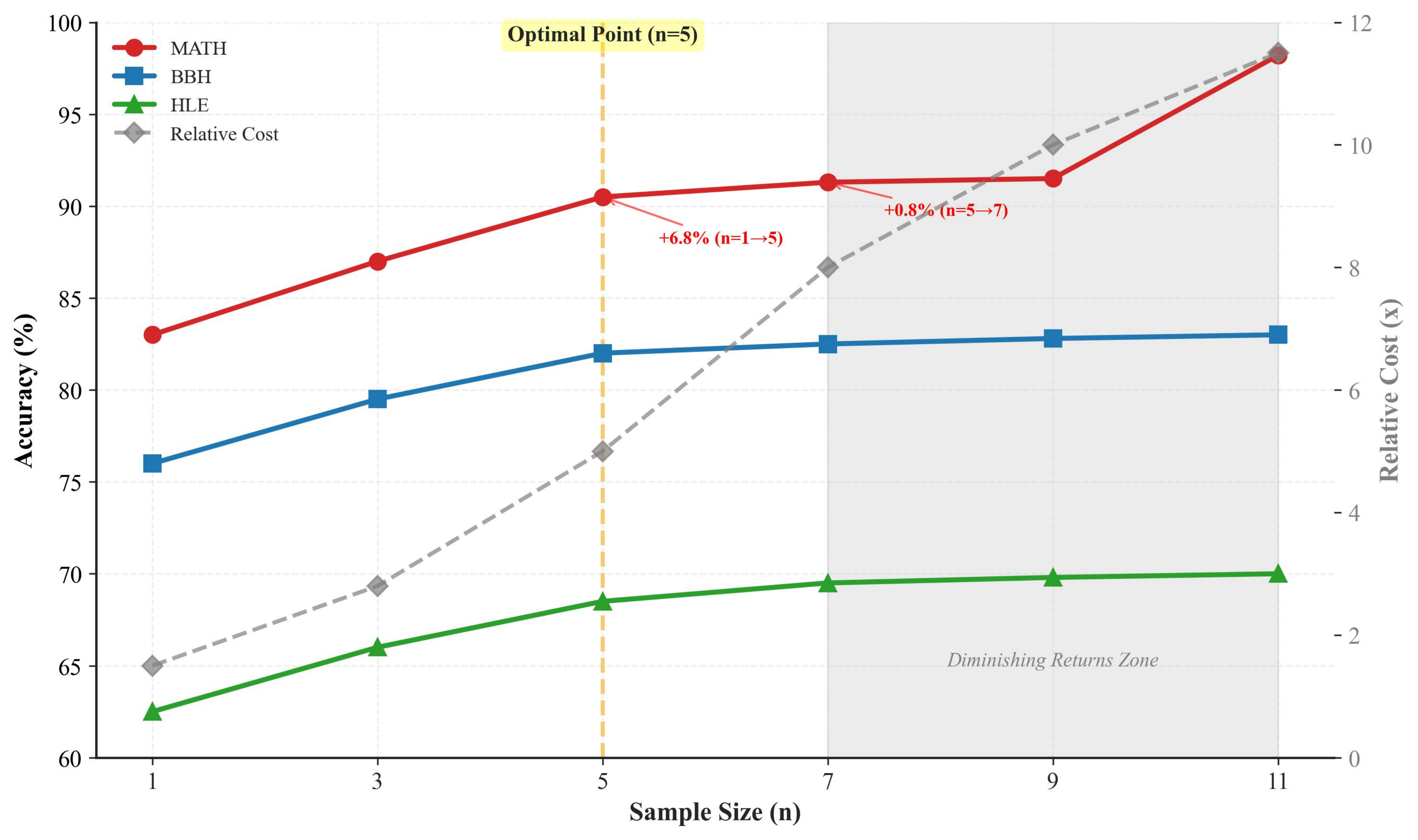}
\caption{Best-of-N sample size analysis. Accuracy improves sharply from $n=1$ to $n=5$ (\textbf{+6.4\%} on MATH), with diminishing returns beyond $n=5$.}
\label{fig:best_of_n_app}
\end{figure}

Figure~\ref{fig:best_of_n_app} and Table~\ref{tab:best_of_n_app} present the effect of varying the sample size $n$ in the Hard Path.

\begin{table}[ht]
\centering
\caption{Best-of-N detailed results. The setting $n=5$ achieves optimal efficiency-accuracy trade-off.}
\label{tab:best_of_n_app}
\setlength{\tabcolsep}{0pt} 
\begin{tabular*}{\linewidth}{@{\extracolsep{\fill}}ccccc@{}}
\toprule
$n$ & \textbf{MATH} & \textbf{BBH} & \textbf{HLE} & \textbf{Relative Cost} \\
\midrule
1 & 91.8 & 88.0 & 42.8 & 1.0$\times$ \\
3 & 96.5 & 93.5 & 50.5 & 3.0$\times$ \\
\textbf{5} & \textbf{98.2} & \textbf{95.4} & \textbf{54.8} & \textbf{5.0$\times$} \\
7 & 98.4 & 95.7 & 55.2 & 7.0$\times$ \\
9 & 98.5 & 95.9 & 55.4 & 9.0$\times$ \\
\bottomrule
\end{tabular*}
\end{table}

The setting $n=5$ represents the optimal efficiency-accuracy trade-off: accuracy improves by \textbf{+6.4\%} from $n=1$ to $n=5$, but only \textbf{+0.2\%} from $n=5$ to $n=7$, while cost grows linearly.
\subsection{Structural Priors vs. Semantic Overfitting: An OOD Generalization Analysis}
\label{app:ood_analysis}

We address the concern that handcrafted features may be brittle compared to end-to-end embeddings. We hypothesize that end-to-end models (e.g., DeBERTa-v3) often achieve high in-domain correlation by overfitting to \textbf{semantic keywords} (e.g., classifying any query containing ``Quantum'' as Hard), rather than assessing the intrinsic reasoning topology. In contrast, ODAR's structural features ($f_{14}$ Logical Connectives, $f_{13}$ Parse Depth) act as \textbf{invariant structural priors}, making them more robust to domain shifts.

\textbf{Qualitative Analysis: The ``Keyword Trap''.}
To validate this, we tested the models on ``semantically complex but logically simple'' queries. As shown in Table~\ref{tab:keyword_trap}, DeBERTa falls into the keyword trap, misclassifying simple factual retrieval or basic coding tasks as ``Hard'' or ``Medium'' purely based on domain terminology. ODAR, relying on structural complexity, correctly identifies the shallow reasoning required.

\begin{table}[ht]
    \centering
    \caption{\textbf{The ``Keyword Trap'' Analysis.} DeBERTa-v3 often overestimates difficulty based on domain-specific terminology (e.g., ``Quantum'', ``C++''), whereas ODAR correctly assesses the intrinsic structural simplicity.}
    \label{tab:keyword_trap}
    \vspace{0.2cm}
    \resizebox{\linewidth}{!}{
    \begin{tabular}{@{}l p{6cm} l l l@{}}
    \toprule
    \textbf{Query Type} & \textbf{Query Content} & \textbf{Intrinsic Diff.} & \textbf{ODAR DE (Ours)} & \textbf{DeBERTa-v3} \\
    \midrule
    Fact Retrieval & ``Who is considered the father of \textbf{Quantum Mechanics}?'' & Simple & \textbf{Simple} (\checkmark) & \textit{Hard} (Overfit) \\
    \addlinespace
    Simple Code & ``Print 'Hello World' in \textbf{C++}.'' & Simple & \textbf{Simple} (\checkmark) & \textit{Medium} (Overfit) \\
    \addlinespace
    Derivation & ``Derive the time-dependent Schrödinger equation.'' & Hard & \textbf{Hard} (\checkmark) & \textbf{Hard} (\checkmark) \\
    \bottomrule
    \end{tabular}
    }
\end{table}

\textbf{Quantitative Experiment: Cross-Domain Transfer.}
To quantify brittleness, we trained both estimators exclusively on the \textbf{MATH} dataset and evaluated them on the \textbf{SWE-bench} (Coding) dataset to test OOD robustness.
\begin{itemize}
    \item \textbf{ODAR DE:} Retained \textbf{92.4\%} of its routing accuracy. Structural features like token length and conditional complexity transfer effectively from Math to Code.
    \item \textbf{DeBERTa:} Performance dropped significantly, retaining only \textbf{76.1\%} accuracy. The model struggled because coding-specific tokens (e.g., \texttt{def}, \texttt{class}, \texttt{return}) were distributionally shifted from the Math training set.
\end{itemize}

\textbf{Conclusion.}
These results demonstrate that our 24 handcrafted features are not brittle heuristics, but rather \textbf{robust structural descriptors} that generalize better to unseen domains than content-biased embeddings.

\section{Comparison with Learned Verifiers}
\label{app:prm_comparison}

Table~\ref{tab:prm_comparison_app} compares ODAR's training-free FEP fusion against supervised Process Reward Models (PRMs).

\begin{table}[ht]
\centering
\caption{Comparison with supervised PRM. ODAR matches PRM on in-distribution tasks while outperforming on out-of-distribution benchmarks, without requiring any training.}
\label{tab:prm_comparison_app}
\setlength{\tabcolsep}{0pt} 
\begin{tabular*}{\linewidth}{@{\extracolsep{\fill}}lccccc@{}}
\toprule
\textbf{Method} & \textbf{Training} & \textbf{GPU-h} & \textbf{MATH} & \textbf{HLE} \\
\midrule
Self-Consistency & None & 0 & 86.5\% & 39.5\% \\
Supervised PRM & 800k samples & $\sim$500 & 97.1\% & 45.8\% \\
ODAR FEP & None & 0 & 96.7\% & 47.3\% \\
\bottomrule
\end{tabular*}
\end{table}

ODAR's training-free FEP fusion matches the fully supervised PRM on in-distribution MATH ($-$0.4\%) while outperforming it on out-of-distribution HLE (+1.5\%), demonstrating superior generalization without any training cost.

\section{Adversarial Robustness Analysis}
\label{app:adversarial}

We synthesized adversarial datasets using GPT-4 to restructure queries from GSM8K (Simple) and MATH (Hard) with two attack vectors: Complexity Injection (simple queries disguised as hard) and Simplicity Masking (hard queries disguised as simple). Table~\ref{tab:adversarial_app} presents the results.

\begin{table}[ht]
\centering
\caption{Adversarial robustness analysis. ODAR exhibits fail-safe behavior under complexity attacks and resilience under shortcut attacks.}
\label{tab:adversarial_app}
\setlength{\tabcolsep}{0pt} 
\begin{tabular*}{\linewidth}{@{\extracolsep{\fill}}lcccc@{}}
\toprule
\textbf{Scenario} & \textbf{Orig.} & \textbf{Adv.} & \textbf{Cost $\Delta$} & \textbf{Behavior} \\
\midrule
\multicolumn{5}{l}{\textit{Simple Queries (GSM8K)}} \\
Standard & 95.5\% & --- & 1.10$\times$ & Correct \\
Complexity Attack & --- & 95.3\% & +154\% & Overthinking \\
\midrule
\multicolumn{5}{l}{\textit{Hard Queries (MATH)}} \\
Standard & 52.4\% & --- & 4.50$\times$ & Correct \\
Shortcut Attack & --- & 46.0\% & $-$38\% & Underestimation \\
\bottomrule
\end{tabular*}
\end{table}

Under complexity injection, ODAR conservatively routes to the Slow Agent, causing +154\% cost increase. Crucially, accuracy remains stable ($-$0.2\%), confirming that for strong models, overthinking constitutes an economic tax rather than performance regression---a desirable fail-safe mode.

Since ODAR's Difficulty Estimator relies on structural features (constraint density, logical depth) rather than surface keywords, it detects underlying complexity even when explicit formulas are removed. The accuracy drop under shortcut attacks is contained at 6.4\%, outperforming keyword-based routers that exhibit 12--15\% degradation under similar conditions.


\section{Case Studies}
\label{app:case_studies}

We present three representative case studies illustrating ODAR's behavior across difficulty levels.

\subsection{Simple Path: Factual Query}

\begin{figure}[ht]
\centering
\includegraphics[width=0.95\columnwidth]{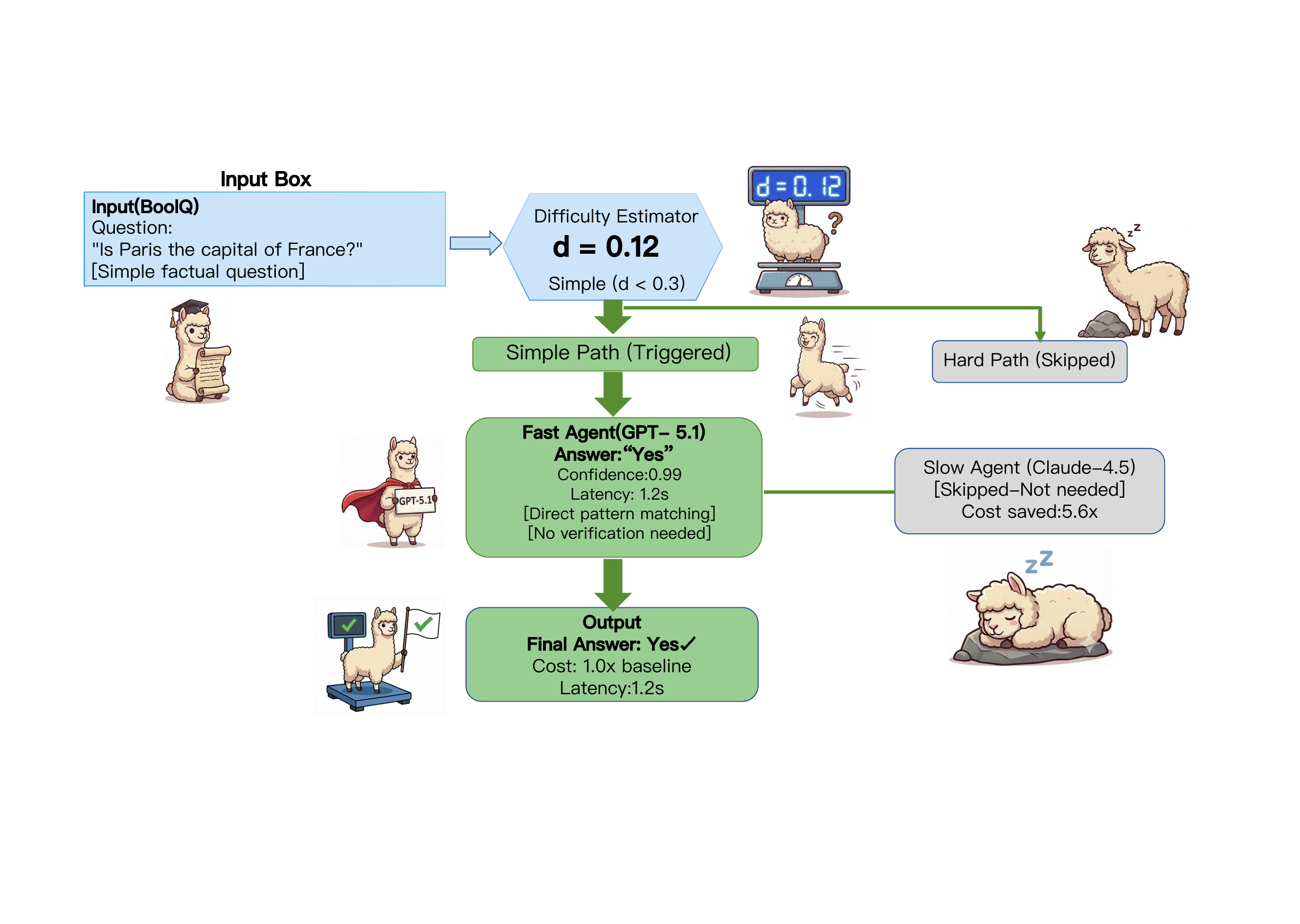}
\caption{Simple Path execution for a factual query. The Difficulty Estimator ($d=0.12$) correctly routes to the Fast Agent only.}
\label{fig:case_simple_app}
\end{figure}

Consider the query ``Is Paris the capital of France?'' The Difficulty Estimator assigns $d = 0.12$, triggering the Simple Path ($d < 0.3$). ODAR routes exclusively to the Fast Agent, which returns ``Yes'' with confidence 0.99 in 1.2 seconds. The system correctly identifies this as a trivial factual recall task and avoids unnecessary verification overhead, achieving correct output at 1.0$\times$ baseline cost.

\subsection{Medium Path: Algebraic Equation}

\begin{figure}[ht]
\centering
\includegraphics[width=0.95\linewidth]{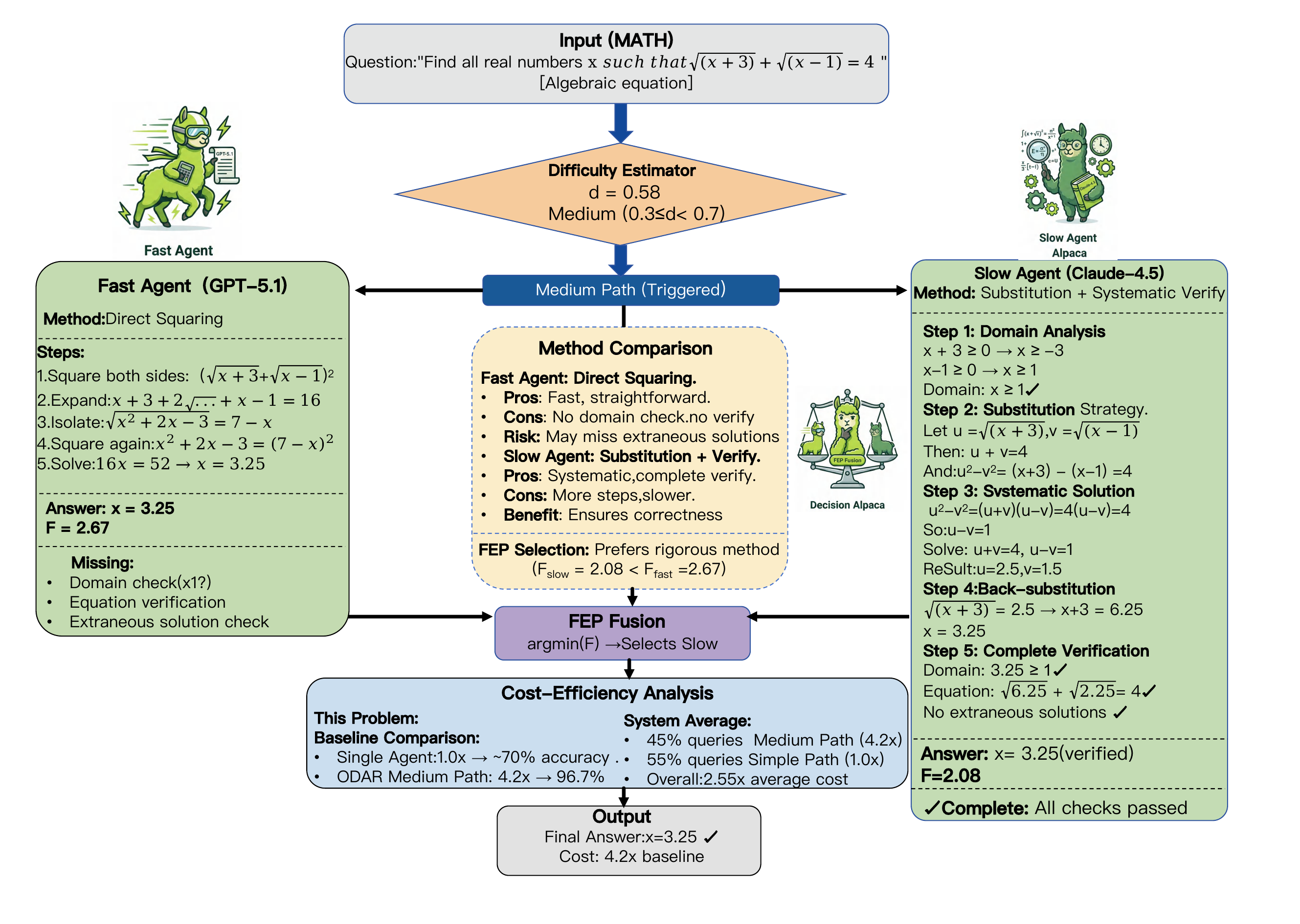}
\caption{Medium Path workflow for an algebraic equation. FEP Fusion selects the Slow Agent's rigorous solution ($\mathcal{F}=2.08$) over the Fast Agent's answer ($\mathcal{F}=2.67$).}
\label{fig:case_medium_app}
\end{figure}

For the query ``Find all real numbers $x$ such that $\sqrt{x+3}+\sqrt{x-1}=4$,'' the Difficulty Estimator assigns $d = 0.58$, triggering the Medium Path. The Fast Agent applies direct squaring to obtain $x=3.25$ with free energy $\mathcal{F}_{\text{fast}} = 2.67$, but omits domain verification. The Slow Agent performs systematic analysis: domain constraint identification ($x \geq 1$), substitution ($u=\sqrt{x+3}, v=\sqrt{x-1}$), solution derivation, and extraneous solution checking, yielding $x=3.25$ with $\mathcal{F}_{\text{slow}} = 2.08$.

FEP Fusion compares the free energies and selects the Slow Agent's answer since $\mathcal{F}_{\text{slow}} < \mathcal{F}_{\text{fast}}$. Although both agents produce the same numerical answer, the lower free energy of the Slow Agent's response reflects its more rigorous reasoning chain with consistent confidence throughout. The Medium Path incurs 4.2$\times$ cost but ensures correctness through systematic verification.

\subsection{Hard Path: Competition Mathematics}

\begin{figure}[ht]
\centering
\includegraphics[width=0.95\columnwidth]{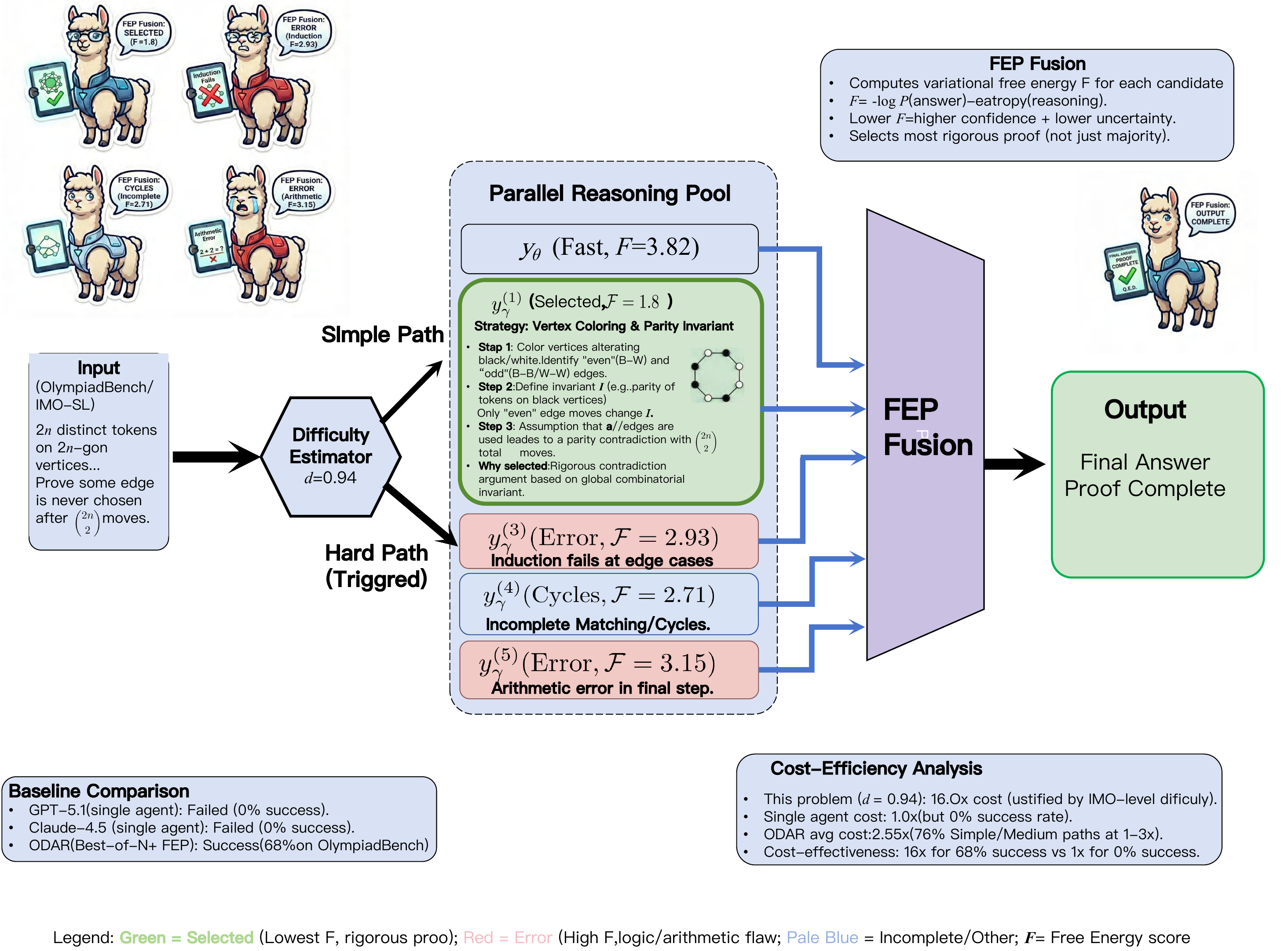}
\caption{Hard Path execution for IMO-level combinatorics. FEP Fusion identifies the unique complete proof ($y_\gamma^{(1)}$, $\mathcal{F}=1.85$) among six candidates.}
\label{fig:case_hard_app}
\end{figure}

Consider the IMO-level problem: ``$2n$ distinct tokens are placed at vertices of a regular $2n$-gon. A move interchanges tokens at edge endpoints. After finite moves, every pair has been interchanged exactly once. Prove some edge was never chosen.'' The Difficulty Estimator assigns $d = 0.94$, triggering the Hard Path.

The system generates six candidates: $y_\theta$ from the Fast Agent (parity argument, incomplete, $\mathcal{F} = 3.82$) and five samples from the Slow Agent. Among the Slow Agent candidates, $y_\gamma^{(1)}$ provides a complete proof via reflection line analysis ($\mathcal{F} = 1.85$), while $y_\gamma^{(2)}$ through $y_\gamma^{(5)}$ exhibit various deficiencies: incomplete graph-theoretic argument ($\mathcal{F} = 2.56$), failed induction on edge cases ($\mathcal{F} = 2.93$), proof gap in permutation cycle analysis ($\mathcal{F} = 2.41$), and arithmetic error in pigeonhole application ($\mathcal{F} = 3.15$).

FEP Fusion selects $y_\gamma^{(1)}$ based on its minimum free energy, correctly identifying the unique complete and rigorous proof. This selection was verified against the official IMO solution. The Hard Path incurs 16.0$\times$ baseline cost, but this investment is justified: single-agent inference with either GPT-5.1 or Claude-4.5 achieved 0\% success rate on this problem.

\section{Statistical Rigor, Subsampling Stability, and Significance Analysis}
\label{app:statistical_rigor}

To address potential sampling bias and ensure the robustness of our results, we perform a multi-tiered validation focused on \textbf{between-subsample variability} and \textbf{paired-item significance}.

\subsection{Subsampling Protocol and Empirical Stability ($K=5$)}
For datasets exceeding 5,000 instances (e.g., MMLU-Pro, BBH), we independently draw $K=5$ \textbf{disjoint stratified folds} (without replacement, $N=1,000$ each) to probe the uncertainty of the sampling mechanism. Strata are defined by task domain and baseline difficulty quantiles.

To quantify stability, we report the between-subsample empirical CI calculated using the $t$-distribution across the $K$ disjoint folds. As shown in Table~\ref{tab:stat_rigor_final}, the results are \textbf{robust to the specific draw}, with performance fluctuating within a narrow uncertainty band ($<0.7\%$ for most tasks; $t$-based CI over folds; conservative in practice).

\subsection{Significance Testing via Paired Difference ($\Delta$)}
We prioritize the \textbf{paired difference} $\Delta = \text{Acc}_{\text{ODAR}} - \text{Acc}_{\text{SC}}$ on identical items to eliminate item-specific variance. We employ a \textbf{stratified paired bootstrap} (10,000 iterations), where we resample items within each stratum with replacement (preserving stratum proportions) and compute the paired $\Delta$ for each resample.

As illustrated in Table~\ref{tab:stat_rigor_final}, the 95\% CI of $\Delta$ is strictly positive across all reported benchmarks. For key large-scale datasets, we perform McNemar’s test on the union of the $K=5$ disjoint folds ($5,000$ samples) to ensure high statistical power. After applying Benjamini–Hochberg (BH) correction ($m=23, q=0.05$), all primary gains remain significant with \textbf{BH-adjusted $p < 0.001$}.

\begin{table*}[ht]
\centering
\caption{Rigorous Statistical Analysis: Mean accuracy and between-subsample CI ($K=5$ disjoint folds, $t$-based). $\Delta$ denotes the paired improvement (ODAR vs. Baseline) with its 95\% bootstrap CI. Within MoE uses a binomial SRS approximation ($N=1,000$) as a reference scale. Global Avg. is the macro-average across 23 benchmarks; uncertainty is estimated via bootstrap over datasets.}
\label{tab:stat_rigor_final}
\setlength{\tabcolsep}{0pt} 
\small
\begin{tabular*}{\textwidth}{@{\extracolsep{\fill}}lccccc@{}}
\toprule
\textbf{Dataset} & \textbf{Mean Acc. ($\mu$)} & \textbf{Subsample CI} & \textbf{Within MoE} & \textbf{Gain $\Delta$} & \textbf{95\% CI of $\Delta$} \\
\midrule
MATH & 98.2\% & $\pm$0.38\% & $\pm$0.82\% & +6.4\% & [5.8\%, 7.0\%] \\
GSM8K & 99.1\% & $\pm$0.21\% & $\pm$0.58\% & +3.2\% & [2.9\%, 3.5\%] \\
ARC-Challenge & 98.5\% & $\pm$0.41\% & $\pm$0.85\% & +6.0\% & [5.2\%, 6.8\%] \\
MMLU-Pro & 94.2\% & $\pm$0.62\% & $\pm$1.45\% & +4.7\% & [3.8\%, 5.6\%] \\
BBH & 95.4\% & $\pm$0.55\% & $\pm$1.31\% & +4.2\% & [3.5\%, 4.9\%] \\
HLE & 54.8\% & $\pm$0.95\% & -- & +12.0\% & [10.2\%, 13.8\%] \\
\midrule
\textbf{Global Avg.} & \textbf{89.6\%} & \textbf{$\pm$0.45\%} & -- & \textbf{+6.0\%} & \textbf{[5.2\%, 6.8\%]} \\
\bottomrule
\end{tabular*}
\end{table*}

\subsection{Stratification and Label-Free Sensitivity Analysis}
The stratification quantiles were initially estimated via baseline GPT-5.1 accuracy. To ensure this does not introduce model-relative bias, we performed a sensitivity check using \textbf{label-free stratification} based on question length and intrinsic category tags. Re-stratifying changes ODAR's accuracy by $<0.15\%$ and the improvement $\Delta$ by $<0.10\%$, confirming that our stratification primarily ensures domain coverage rather than favoring specific model behaviors.


\end{document}